\documentclass[a4paper,twocolumn,array]{revtex4}
\usepackage{graphicx}

\begin{document}

\title{Null Dynamical State Models of Human Cognitive Dysfunction}

\author{Michael J. Gagen}

\affiliation{\rm Email: mjgagen@gmail.com}
\date{\today}

\begin{abstract} 
The hard problem in artificial intelligence asks how the shuffling of syntactical symbols in a program can lead to systems which experience semantics and qualia. We address this question in three stages.  First, we introduce a new class of human semantic symbols which appears when unexpected and drastic environmental change causes humans to become surprised, confused, uncertain, and in extreme cases, unresponsive, passive and dysfunctional. For this class of symbols, pre-learned programs become inoperative so these syntactical programs cannot be the source of experienced qualia. Second, we model the dysfunctional human response to a radically changed environment as being the natural response of any learning machine facing novel inputs from well outside its previous training set.  In this situation, learning machines are unable to extract information from their input and will typically enter a dynamical state characterized by null outputs and a lack of response.  This state immediately predicts and explains the characteristics of the semantic experiences of humans in similar circumstances. In the third stage, we consider learning machines trained to implement multiple functions in simple sequential programs using environmental data to specify subroutine names, control flow instructions, memory calls, and so on.  Drastic change in any of these environmental inputs can again lead to inoperative programs.  By examining changes specific to people or locations we can model human cognitive symbols featuring these dependencies, such as attachment and grief. Our approach links known dynamical machines states with human qualia and thus offers new insight into the hard problem of artificial intelligence.  
\end{abstract}

\maketitle

\section{Introduction}

Artificial intelligence seeks to understand how a machine interacting with its environment, processing syntactical symbols, and changing from one causal dynamical state to another, can generate semantic symbols, meaning, intentionality, and understanding. This is the hard problem of artificial intelligence \cite{Chalmers_95_20,Chalmers_96}. There are two contrasting approaches to modeling artificial intelligence, both grounded in work by Turing. In the first approach, Turing defined a universal computer and its set of computable functions which, with the benefit of hindsight, raises the question of whether intelligence is itself a computable function \cite{Turing_1937}.  However, rather than addressing this question, Turing settled for the simpler claim that merely simulating intelligent behaviour was a computable function and proposed the imitation game, now called the Turing Test, as the only practical approach to discerning intelligence whether natural or artificial \cite{Turing_1950}. 

The hypothesis that intelligence itself is computable was most strongly presented in 1976 with the physical symbol systems hypothesis (PSSH) holding that syntactical symbol manipulations are necessary and sufficient to explain and reproduce human and machine intelligence \cite{Newell_1976_11}. This approach posits a system of symbol structures, each consisting of many atomic symbols which are processed according to a set of creation, reproduction, modification and destruction rules. This system inhabits a world of external objects, where symbols can ``designate'' an object if that symbol can influence or be influenced by the object, and where symbols are ``interpreted'' if they designate a process which the system can carry out. For many decades, this approach has been interpreted in extremely narrow and limited terms.  However, broadly interpreted, the PSSH subsumes many subsequent research programs \cite{Newell_88,Nilsson_07_9}. An invocative and poetic treatment of the potential of a symbolic approach sees future conscious programs as being based on active symbols equivalent to self-contained programs possessing multiple simultaneous interpretations on many different levels, with symbols causally linked with internal and external environments, and eventually combining to form a self-referential model of consciousness \cite{Hofstadter_79}. 

The PSSH approach has been critiqued by many authors.  In 1978, it was argued that however accurately a syntactical symbol system modeled human physiology, that system (a computer or the Chinese nation) would never feel pain \cite{Dennett_78_41,Block_78}.  Just afterwards, Searle introduced a human into a syntactical symbol system, the Chinese Room, to show that no amount of syntactical symbol processing would ever lead to semantic meaning and qualia as experienced in human cognition \cite{Searle_80_41}. In the subsequent widespread and ongoing debate, some have argued that consciousness doesn't exist \cite{Rey_97_46,Dennett_91}, or that consciousness can only be explained by extensions to currently understood physics \cite{Chalmers_96,Penrose_89}. 

The systems reply to the Chinese Room holds that these perceived limitations reflect a failure of human intuition about just what syntactical symbolic programs can achieve. For instance, simple dynamical systems with only three degrees of freedom \cite{Moore_90_23}, and simple train networks \cite{Chalcraft_94_5}, can implement undecidable computations and hence have totally unpredictable outcomes. In 1988, Boden noted that intuitions about the capacities of computer programs couldn't be trusted, and argued that computer programs are not merely formal systems or syntax alone as they can, given a suitable hardware context, cause procedures to be implemented, which builds or activates symbols, which can alter external objects creating a causal semantics \cite{Boden_90}. Exactly how syntactical symbols grounded in sensory inputs can combine to form higher level semantic symbols is the still open symbol grounding problem\cite{Harnad_90_33,Taddeo_05_41}. It can be argued that these latter statements of the problem essentially reprise the designated and interpreted symbols of the physical symbol systems hypothesis \cite{Newell_1976_11}. Researchers have been building on our understandings to formulate models of human level or artificial general intelligence \cite{Adams_12_25,Goertzel_07,Nilsson_05_68,Minsky_04_11}. 

Despite our difficulties in understanding the hard problem of artificial intelligence, there has been considerable progress lately in artificial intelligence systems.  Massive data tables and number crunching allowed Deep Blue to be crowned as world chess champion \cite{Campbell_02_57}, while Watson employed big data and natural language processing to defeat human Jeapardy champions \cite{Ferrucci_10_59}. Deep learning neural networks \cite{Hinton_06_15,Deng_13_197,LeCun_15_436} and deep neural networks undertaking partially unsupervised and reinforcement learning have seen computers master video games such as ATARI \cite{Mnih_15_529}, and seen AlphaGo, AlphaGo Zero, and AlphaZero achieve superhuman performance at Go, Chess, and Shogi \cite{Silver_16_484,Silver_2017_354,Silver_2017_01815}. Neuroscience is increasingly optimistic about locating the neural correlates of consciousness \cite{Crick_1994,Koch_2004}, while the growing power of computers is being exploited to implement cortical simulators \cite{Hawkins_2004,Merolla_14_668} leading to whole brain emulators for C. elegans \cite{Towlson_2013_6380}, fly \cite{Givon_16_fly}, mouse \cite{Oh_14_207}, and eventually it is hoped, human.  Interestingly, the ability to visualize reconstructed human visual brain experiences has been demonstrated \cite{Nishimoto_11_1641}.  

However, despite all the progress being made, few would argue that these programs possess understanding or experience qualia. Further, the size and complexity of these programs suggests that any putative artificial intelligence program able to mimic a full range of human capacities will be both enormous and complex. In turn, the need for complex programs to model artificial intelligence implies  that there is nothing to be learned from small systems lacking size or complexity or code.  

In this paper, we wish to present a new approach to artificial intelligence which exploits small systems lacking both code and capacity, and yet which are able to generate dynamical states whose characteristics predict and explain the semantic experiences of a particular class of human semantic symbols.  In this new class, humans become cognitively dysfunctional when faced with radical environmental change as in a disaster.  Because humans are largely dysfunctional, then an accurate model of human cognition in these circumstances does not require large and complex code.  For this class of semantic symbols, we claim that simple learning machine models can predict and explain aspects of the human semantic experience.  We make no claims that our simple models will themselves have semantic experiences.  We turn to consider how this approach will be implemented here. 

In this paper, we introduce a new class of cognitive semantic symbols which humans typically experience when faced with radical and unexpected environmental change as in a disaster. This class is introduced in Sect. \ref{sect_new_class_cognitive_symbols}. In a disaster, humans will continue to experience semantics and qualia but many of them will become cognitively unaware, passive, unmotivated, and unresponsive. They will be unable to access any of their pre-existing or pre-learned functional capacities so, as noted above, these pre-learned capacities cannot be the source of human qualia and do not need to be included in our small models. Our approach breaks the linkage from syntax to semantics.   

It is also relatively straightforward to model the dysfunctional human response to a disaster using a learning machine.  In particular, every learning machine presented with input data well outside its previous training set will generally become dysfunctional---an example of ``garbage in, garbage out''.  More interestingly, a sufficiently complicated learning machine will generally enter a particular dynamical state when presented with novel inputs.  Because the machine can't extract information from the unknown input, the machine ends up processing zero information and so exhibits a ``null'' dynamical state. This null dynamical state serves as a close model of the unresponsive cognitive state of humans in similar circumstances.  In Sect. \ref{sect_confused_Chinese_Room}, we show that the characteristics of this null dynamical state can be used to predict and explain the characteristics of human semantic symbols in similar circumstances.  This learning machine model is then used to explain the shocked and confused behaviour of Searle's Chinese Room when placed in a disaster scenario.  (Because the Room is an accurate simulation of a human, it will respond like a human and appear to be passive and unresponsive.)  The objectively observable dynamical null state of the Room will predict and explain the subjective experience (if any) of the Room and of humans in disaster situations.  We will finish this section by discussing how a dynamical null state model addresses the subjective-objective explanatory gap between a computational state and semantic experiences, and the properties of qualia as experienced by humans.

In Sect. \ref{sect_learning_machine_null_states}, we provide a mathematical specification of the dynamical null state typically exhibited by learning machines when presented with data well outside their training sets. We focus principally on artificial neural networks as their dynamics are reasonably well understood.  The result is a mathematically specified dynamical null state whose characteristics predict and explain the semantic cognitive experiences of humans in similar situations.

We apply these dynamical null states to model human semantic symbols in Sect. \ref{sect_attachment_grief_spoof} where we ask what happens when a large percentage (80\% or so) of a learning machine's functional capacities have been learned in the presence of a single teacher (Alice say) and at a single location (Alice's laboratory). The lack of variation in the environment during the learning period makes it likely that environmental information unique to the teacher or to the location will be deeply embedded within the machine's learned functions---rather than learning a generic ``Say Good Morning'' function, the machine might learn a specific ``Say Good Morning to Alice'' function.  Consequently, if Alice is absent and Bob says ``Good Morning'' to the machine, then it will be unable to respond as it has no learned function specific to Bob.  In our approach, we consider how a learning machine can build environmental location and person dependencies into its functionality, so that drastic changes to the environment can cause dysfunction.  We apply this approach to model a number of human cognitive semantic symbols.  

Finally, a concluding summary appears in Sect. \ref{sect_conclusion}.

\section{A New Class of Semantic Symbols: Human Disaster Response}
\label{sect_new_class_cognitive_symbols}

We have no idea how to even begin coding a program to understand, to intend, or, for instance,  to experience ``redness'' on observing red photons. However, it might be possible to make progress by considering a new class of human cognitive symbols which arise when humans face unexpected radical environmental change and generate qualia like surprise and confusion.  Because the changed environment is unexpected, we claim that the qualia don't result from pre-existing programs, but arise from the altered dynamical state of the underlying cognitive architecture.   

The responses of people to unexpected drastic environmental change is often not as cool, calm and collected as we would hope, or as panicked as we might suspect.  Disaster response organizations have labeled many common beliefs about how people respond as myths.  In particular, ``Most disaster victims are psychologically resilient and engage in socially integrative---rather than destructive---responses'' \cite{Lindell_2006}.  However, a small percentage of the population (of order 14\%) can display disaster shock and respond with docility, disoriented thinking, apathy, confusion and disbelief when facing ``sudden onset, low forewarning events involving widespread physical destruction''  \cite{Lindell_2006}.

Anthropological studies of disasters led Wallace to summarize a typical human response to ``unexpected, sudden and overpowering trauma'' as the disaster syndrome \cite{Wallace_1956_1,Wallace_1956_3,Wallace_2003}.  A ``large proportion of persons in the impact area'' (up to 33\%) may ``appear to the observer to be `dazed,' `stunned,' `apathetic,' `passive,' `immobile,' or `aimlessly puttering around'.''  Wallace concluded that the ``determinants of the syndrome are, in my opinion, not primarily physical injuries, physical shock, \dots they are psychological'' as a person displays the ``syndrome whether or not he has been injured.'' Wallace felt the ``precipitating factor in the disaster syndrome seems to be ``the perception that \dots practically the entire visible community is in ruins.'' In response to this realization, victims exhibit ``withdrawal from perceptual contact with this grim reality and regression to an almost infantile level of adaptive behavior characterized by random movement, relative incapacity to evaluate danger or to institute protective action, inability to concentrate attention, to remember, or to follow instructions.'' \cite{Wallace_1956_1,Wallace_1956_3,Wallace_2003}.  

The disaster syndrome, as described by Wallace, is nowadays seen to be part of a broader suite including disaster response, psychological shock and physiological conservation-withdrawal syndrome \cite{Valent_2000_706}.  Victims of disaster can exhibit shock, stupor, being dazed, stunned and numb, and can exhibit immobility, quiescence and unresponsiveness, constricted attention and detachment.

In this paper, we will be emphasizing the importance of psychological factors (rather than those caused by physical injury or physiological hormonal responses) in sudden disasters when victims glimpse ``a part of the human condition that most of us never see'' \cite{Ripley_2009}. These factors are clearly indicated by the first-person report of an evacuee from the towers on the ``9/11'' attacks.
Elia Zede\~{n}o felt the plane impact the tower, and even after a colleague screamed at her to get out, still found herself collecting belongings and wandering in circles at her desk, and feeling in a trance. Roughly 40\% of survivors gathered belongings before evacuating.  Zede\~{n}o's later reported response to seeing bodies lying motionless outside the building lobby illustrates  ``how the human mind processes overwhelming peril'': 
\begin{quote}   
I'm slowing down because I'm starting to realize I'm not just looking at debris.  My mind says, ``It's the wrong color.'' That was the first thing.  Then I start saying, ``It's the wrong shape.'' Over and over in my mind. \dots It was like I was trying to keep the information out.  My eyes were not allowing me to understand.  I couldn't afford it.  \dots Then when I finally realized what it meant to see the wrong color, the wrong shape, that's when I realized, I'm seeing bodies.  That's when I froze.
\end{quote}
``Zede\~{n}o went temporarily blind at that moment,'' and felt numb, and had to be led outside the building. As the tower came down, her vision returned when needed, and she survived  \cite{Ripley_2009}.  

Similar responses to disaster have been collated by Leach.  An ``engine fire aboard
a Boeing-737 airliner at Manchester airport in 1985 that resulted in 55 deaths
found some passengers sitting immobile in their seats until overtaken by smoke and
toxic fumes'' \cite{Leach_11_26}.  Similar passivity substantially increased the death toll in the Piper Alpha oil platform fire in which ``A large number of people apparently made no attempt to leave the accommodation'' \cite{Cullen_1990}.  Even prepared people can suffer unexpected passivity under stress---11\% of parachuting fatalities result from ``no-pull'' events where after failure of the main chute the reserve chute is not pulled \cite{Griffith_02_10}.  Leach concluded that victim passivity results from a ``difficulty in integrating information from the new survival environment, through multimodal systems, to information stored in long-term memory'' so that ``victims perish unnecessarily because the threat environment restricts both the storage and processing capacities of working memory coupled with a form of temporary, environmentally induced dysexecutive syndrome'' \cite{Leach_11_26}.  After subsequent investigations, Leach found that victim immobility or freezing is a ``common response to unfolding emergencies'' resulting from an ``impaired response that delays evacuation, establishing a closed-loop process that leads to fatalities in otherwise survivable situations'' \cite{Leach_04_539}. Examples cited included 
\begin{itemize}
\item volunteers who were ``behaviorally inactive'' in practice airline evacuations, 
\item passengers in the Manchester crash ``were seen to remain in their seats until they became engulfed in flames,'' 
\item survivors of the Piper Alpha oil rig fire reported that victims could not be made to move and ``just slumped down,'' 
\item some victims of the Estonia ferry sinking were ``passive and stiff despite reasonable possibilities for escaping,'' while many apparently perished because they simply did nothing to save themselves.  Survivors reported that victims ``were standing still apparently in shock,'' or ``some paralyzed and exhausted passengers were standing on the staircase'' or were ``sitting in corners,
incapable of doing anything,''  while ``people were beyond reach and did not react when other passengers tried to guide them, not even when they used force or shouted at them.''  One survivor reported that ``I didn't think. Shock is so disorienting it doesn't allow us to think clearly. \dots People just sitting in complete shock and me not understanding why they're not doing something to help themselves.
They just sat there and being swamped by the water when it came in.''  Another survivor reported seeing ``around 10 persons lying on the deck near the bulkhead. They seemed apathetic and he threw life jackets to them. He did not see them react or put on the lifejackets.''
\end{itemize}
Leach concluded that about 10-15\% of people in a disaster will remain relatively calm, able to collect their thoughts and their judgment and reasoning abilities will remain relatively unimpaired. A second group of about 75\% will ``be stunned and bewildered, showing impaired reasoning and sluggish thinking. They will behave in a reflexive, almost automatic manner.'' A final group of about 10-15\% of the population ``will tend to show a high degree of counterproductive
behavior adding to their danger, such as uncontrolled weeping, confusion, screaming, and paralyzing anxiety.'' \cite{Leach_04_539}. In seeking to explain these behaviours, Leach proposed a model in which human working memory has limited capacity and a maximum rate at which novel information can be processed. This ``helps to explain the slowing or absence of response during the critical impact phase of a disaster.''  In addition, Leach proposes that over time people learn complex behaviours as pre-learned behavioural responses or schemas.  In an emergency however, if no pre-learned schema exists then a new temporary schema has to be created and this can take longer than the duration of the emergency. The ``result is that no behavioral schema will be triggered from the schemata database and no temporary schema can be created within the time available. This produces a cognitively induced paralysis or `freezing' behavior'' \cite{Leach_04_539}.  Leach later developed theoretical models to explain this maladaptive behaviour \cite{Leach_12_1152}.  

These human responses to disaster have been widely studied with the view of finding an adaptive explanation for this disaster response \cite{Valent_2000_706,Barlow_2002,Baldwin_2013_1549}.  For instance, an adaptive acute stress response to danger sees animals first freezing to hide and assess the situation, then fleeing danger if possible, fighting if not, and fainting to ``play dead'' and exhibit tonic immobility if caught by a predator \cite{Bracha_2004_679}.  These added fainting stages feature a dissociative ``shut-down'' exhibiting a ``partial or complete loss of the normal integration, \dots immediate sensations, and control of bodily movements'' and losses of function such as ``hysterical blindness'' with no evident organic basis \cite{Schauer_2010_109}.  Tonic immobility has been linked to rape-induced paralysis wherein a high percentage of victims feel paralyzed and unable to act despite no loss of consciousness \cite{Heidt_2005_1157}.  Experiments into tonic immobility are ongoing \cite{Schmidt_2008_292,Volchan_2011_13}.

While few people will experience a major disaster and the associated response, many have experience a range of lesser events involving an unexpectedly changed environment.  These lesser shocks lie on a spectrum and all involve a response to sudden environmental change.  These changes can range in severity from grief and bereavement, to shame and humiliation, to being simply surprised and startled.  On the more innocuous end of the spectrum, surprise and startle become pleasurable and sought after as in amusement parks, and can form the basis of humour which depends on a punch line introducing sudden and unexpected setting changes in a joke.  

In more detail, grief and bereavement can be a debilitating response to the deaths of loved ones.  Responses to bereavement have been described as an ``assault,'' a ``blow to the face, head, or guts,'' of being ``overwhelmed'', of ``not being able to take it, the world shattering around one, and sinking into a black hole,'' with defenses including ``disbelief, and dissociation,'' and of feeling numb \cite{Valent_2000_706}. More general grief symptoms can include fear of losing one's mind, hallucinations, feeling hopeless, fatigue and loss of interest and ability to concentrate \cite{Clayton_2000_304}.  About 10-20\% of people can suffer complicated grief and be overwhelmed by ``thoughts of the deceased, disbelief, feeling stunned, and lack of acceptance of the death; \dots with enduring functional impairments'' \cite{Horowitz_1997_904}.  Sufferers experience a ``persistent disturbing sense of disbelief regarding the death'' and a ``resistance to acceptance of the painful reality'' \cite{Shear_2005_253}. 

In everyday life, an acute stress response can be caused in situations featuring shame, humiliation and embarrassment.  Victims of humiliation can report that ``they felt wiped out, helpless, confused, sick in the gut, paralyzed, or filled with rage,'' or had been ``stabbed in the heart, or hit in the solar plexus,'' so much so that they ``wished they could disappear'' \cite{Klein_91_93}.  Who hasn't experienced the mind-numbing effects of being the unwanted center of attention in a humiliating situation.  

The sensation of surprise results from a ``mismatch between one’s mental expectations and perceptions of one’s environment'' \cite{Rivera_2014} which then causes something similar to the flight or fight reaction with an ``interruption of ongoing information processing and the reallocation of processing resources to the unexpected event'' in order to update the relevant schemas \cite{Reisenzein_2006_295,Rivera_2014}. For instance, pilots can be lulled into such a sense of security because of modern aircraft reliability that ``when unexpected critical events occur, pilots are often genuinely surprised and don't have readily accessible action plans on how to deal with them'' \cite{Martin_2014}.  These surprise events can lead to a startle response and even to acute stress where pilots have sometimes ``taken no action at all'' because of a ``significant impairment to both cognitive and psychomotor performance'' \cite{Martin_2014}.  An investigation into these responses showed a third of pilots in flight simulator experiments exhibited ``pathological reactions'' when startled, and they froze and were unable to respond to unfolding events \cite{Martin_2014}. However, as noted previously, surprise and startle can sometimes be pleasurable and sought after as when we seek it out in amusement parks, or arrange to surprise friends in order to please them---many adults have played ``peek-a-boo'' with a child and enjoyed their happy response.  
  
And everyone enjoys the pleasure of humour which is itself based on the human surprise and startle reactions.   In the incongruity-resolution model for humour, an audience initially interprets a joke using one mental schema which is disrupted when the punchline presents an incongruity whose resolution requires a mental shift and re-evaluation of previously assumed knowledge \cite{Ritchie_1999_78}.  This approach has been widely applied in, for example, children's humour and advertising \cite{Shultz_1972_456,Alden_2000_1}.  Deeper explanations for why these incongruity-resolutions are found pleasant can be based on locating neural correlates in which one neural pattern is unexpectedly replaced by another \cite{Katz_1993_59,Samson_2009_1023}, or broader evolutionary models \cite{Polimeni_2006_347}, or cognitive approaches \cite{Hurley_2011}.

\section{A Shocked and Confused Chinese Room}
\label{sect_confused_Chinese_Room}

When faced with drastic and unexpected environmental change, humans sometimes cannot process novel environmental information, cannot access pre-learned programming, and cannot invent new responses in the time available.  Their cognition can effectively cease and they can become passive, unresponsive and unaware. By design, the Chinese Room will react just like a human in a disaster and so can also become unresponsive.  In this section, we will present a learning machine model of this disaster response in the Chinese Room and in humans.

\subsection{A learning machine model}

In a disaster humans lose access to most pre-existing and prelearned functionalities and yet still experience semantic content, so these prelearned functionalities cannot be the source of human semantic content in a disaster.  Consequently, modeling a human disaster response in an artificial intelligence does not need to include these unused functionalities.  (To illustrate, when modeling how a chess program performs in an altered environment---when put in charge of an autonomous vehicle for instance---we don't need to actually include the thousands of lines of chess playing code that will never be called.)  Similarly, our artificial intelligence model does not need to include any syntactical programs or learned functions that are uncalled in a disaster scenario. A small scale model suffices to model the human disaster response allowing us to break the link between syntactical code and semantic symbols and drastically simplify our approach to artificial intelligence.  

We can further simplify our approach by noting that essentially all sufficiently complicated learning machines will generate a passive and unresponsive state when faced with a novel or radically changed environment well outside their prior training sets. It takes a long time for learning machines to master their training sets, and no learning machine can instantly master a new training set corresponding to a radically altered environment.  Consequently, all learning machines will flounder when faced with novel inputs well outside their existing training set.  In this paper, we are using a broad definition of learning machine as being any machine optimized for a particular environment whether by evolutionary selection, by training and learning, by iterated human design, or by any generalized Darwinian algorithm incorporating replication, variation and selection which increasingly fits the machine to its environment \cite{Holland_1992,Kauffman_1993,Dennett_1995}. Once a learning machine has been fitted to an environment, its training set, then radical changes to that training set will prevent it from extracting information from the new training set.  It will then have nothing to pass to later information processing stages, and the machine will generally output null vectors.  (These claims will be justified in more detail below.)  Hence, a simple learning machine model predicts that radically changed environments will cause a learning machine to continually output null vectors and so exhibit a dynamical state in which it essentially appears ``off''.  The machine itself is ``on'' and fully operational; it is just that it cannot process the information from the radically changed environment. In this dynamical null state, the learning machine becomes passive, unresponsive and ``unaware'' of its surroundings.  It is immediately apparent that the dynamical state of the learning machine in a radically changed environment closely mimics, and predicts and explains the semantic content of human learning machines in similar situations. Emphatically, we are not writing subroutines to simulate passivity, we are finding those circumstances that render all learning machines passive, and using this learning machine response to model the same passivity in human learning machines.  We are using the dynamical state of the machine, its system state, as a predictor of the semantic content experienced by humans in similar situations. This approach supports the systems reply to the Chinese Room argument.  

Another important aspect of our approach is that human brains learn without having external designers optimizing the details of the neural architecture.  For humans, no designer chooses the number of neural modules used, or the number of neural layers and neurons and their activation functions, or the network data representations used to learn a function.  These network architectural choices can radically effect the operation of a neural network---an unstructured learning task such as backing up a lorry \cite{Nguyen_1990_596} can be learned significantly faster and by a smaller network when using external programmers to decompose the task into subproblems \cite{Jenkins_1993_718}. Also, possibly the very ``formation of appropriate representations lies at the heart of human high-level cognitive abilities \dots of understanding how to draw meaning out of the world''  \cite{Chalmers_1992_185}.  Having external designers impose their own high level representations on a network might well prevent it from modeling high level cognitive abilities.  In this paper, we will model learning machines which have trainers but lack external designers.  Our machines will (somehow) determine their own modular structures, decide the number of neural layers and neurons to devote to a task, and formulate their own high level data representations.  If, as is likely, they make mistakes then these mistakes can be used to model the same mistakes made in the development of human neural networks.  If too few neurons are assigned the machine (and the human) will be unable to either learn a task or generalize it, while if too many neurons are assigned then the machine (and human) will overfit to the task and also be unable to generalize. 

For a specific example, consider how a learning machine or a human must learn to acquire data from their environment by developing their own application programming interfaces (APIs)---to read today's temperature the machine or the human might learn to move their visual input system to a certain position to locate a particular rectangular shape and to correlate this mark with that number.  APIs are generally specific to a particular hardware environment implying again that radical environmental change can render APIs inoperative to destroy the machine's input and output channels.  

We will also task our machines with learning multiple functions simultaneously (just as humans do) and with recurrently passing information from one function to another to implement sequential processing.  To accomplish this, our machine must use environmental information to define the equivalents of subroutine names, function names, variable names, memory and data addressing schemes, and the usual control structures like conditional tests and loop counters.  Essentially, our learning machine (and humans) must invent their own new computational architecture and dedicated programming language solely from their interactions with their trainer and their environment.  This is a difficult task, and it is to be expected that environmental data will inappropriately populate all aspects of the learned architecture.  This again means that a radically changed environment can be expected to destroy the learned computer architecture and its programs.  In our approach, a learning machine faced with radical environmental change will be attempting to run barely functional program code and will exhibit passivity, hesitations, repeated loops, glitches, and stutters, just as do humans.  The generated null dynamical state was never learned or pre-programmed, just as with humans, and it is this state that provides our model for the semantic experiences of humans in these situations.

\subsection{Learning Machine Responses to Environmental Change}

A final part of our modeling approach notes that learning machines can respond to radically altered environments in three main ways. First, the learning machine might seek to restore the environment to its original condition.  Second, the learning machine might embark on some lengthy and difficult retraining to fit to the new environment.  Finally, the machine might adopt the easier approach of changing the input data coming from the altered environment back into a form suitable for processing by its original network. In other words, the machine might ``spoof'' or alter its own input data to allow processing of the changed environment by its original network \cite{Dordal_2016}.  (A fourth approach might involve the machine undertaking an automatic ``freeze, flight, fight'' response whenever they are startled by environmental change. We do not include such automatic responses here.)  In this paper, each of these three possible responses to the initial stimulus, the unexpected environmental change, will be used to predict and explain semantic experiences of humans in similar circumstances.

Before considering how the Chinese Room responds to a radically changed environment, it is worthwhile examining some simpler systems.  We will consider model learning machines which receive input data packets from the environment and process and recurrently address them across a network to different functional modules until a sequential program is complete.  

For a first example, consider the post office which is often used as an introduction to packet switched computer networking \cite{Dordal_2016}.  A post office using optical sensors to sort envelopes might well stop working if the spectrum of the room lighting is changed.  The machines will be internally fully functional but unable to work in the changed environment and the system will stop and enter a null dynamical state.  Restarting the system will involve either fixing the room lighting to return the environment to its original state, or changing the machine's optical sensors to suit the new environment, or by spoofing the old sensors using filters (say) to allow the original machine to process information from the new environment. 

For a second example, consider the packet switched internet which is undergoing a designed shift in addressing protocol from the IPv4 (32 bits: e.g. 172.16.254.1) to the IPv6 Internet Protocol (128 bits: e.g. 2001:0db8:85a3:0000:0000:8a2e:0370:7334) \cite{Dordal_2016}.  Those parts of the internet using the older protocol can only operate in the new environment by either forcing the entire internet to return to the old IPv4 protocol, or by updating itself to the new IPv6 protocol, or by spoofing incoming and outgoing packets by either using dual stack routers capable of handling both protocols or by using network address translation or by tunneling and embedding schemes \cite{Dordal_2016}.  

Our third example considers a cellular signal transduction network which uses many molecular interactions to implement combinatorial logic functions and alter protein production dynamics \cite{Watson_2014}. A change in cell salinity (say) might alter the efficacy of the signaling pathway, and this can be repaired only by either returning salinity to normal, re-evolving the entire pathway to suit the changed salinity, or perhaps by introducing new medicine molecules to alter existing molecular binding sites to suit the changed salinity conditions and restore function.  

For a final example, consider a bureaucracy exchanging memos or information packets across its network.  Every large organization in history has faced the challenge of authenticating its memos \cite{Schofield_2015} and detecting forgeries \cite{Hector_1959}, including in modern internet based commerce \cite{Balloon_2001_905}.  A forged information packet uses spoofed authentication codes to subvert the normal operation of the network for illicit gain.  An interesting example occurred in the last days of the Ching Dynasty \cite{Chang_2013} where the mother of the new child Emperor fooled the council of regents into passing a law requiring that all future official edicts needed a stamp held only by the mother.  After that, the regents could not issue official edicts to rescind that previous decision and return the environment to its initial state, could not alter their own state by gaining access to the required stamp, and could not legally spoof or forge the use of the new stamp.  As a result, they lost power.

\subsection{A Shocked and Confused Chinese Room}

Based on the above discussion, it is now possible to consider how an unexpected and radical environmental change might put the Chinese Room \cite{Searle_80_41} or the Chinese Nation \cite{Block_78} or a simulated nervous system \cite{Dennett_78_41} into an unplanned dynamical state characterized by ``null'' outputs.  These systems, by design, will respond as humans do and so will likely exhibit a disaster response. Applying traditional ideas, we might presume that the Room encodes some syntactical code which, when implemented by the man within the fully functional Room, causes the internally fully functional Room to externally simulate dysfunctional cognition. This starkly confirms that merely simulated behaviours carry no implications whatsoever about internal computational states (or experiences). In contrast, our approach seeks to actually duplicate human cognitive dysfunction in a learning machine, and use this to model human cognition. When the Chinese Room faces a disaster, then all pre-existing syntactical code is either irrelevant (it won't be called) or becomes dysfunctional, and the Room likely enters a null dynamical state. The characteristics of this null state can then predict and explain the characteristics of the semantic shock and confusion experienced by humans in similar circumstances.  

It is difficult to be precise, but a short story about this situation might be useful.  We first note that the program encoded in the book in the Room would be designed for a finite set of possible environments (as with humans), and would include only a fraction of the possible Chinese characters (as with most Chinese speakers). Suppose now that the Chinese Room is located within a sinking and burning ship (similar to examples given previously) and that consequently, the input slot for the Chinese characters partially fills with water and smoke, and the input symbols coming through the slot are burnt and smudged and unclear, or are entirely new Chinese characters unknown to the book program. Because the characters cannot be identified, the man in the Room cannot match the characters to the book and he will be left fruitlessly leafing through the book looking for some hint about what symbols to process.  Both the man and the database are still fully functional, and yet no output symbols are being produced.  It is only the alteration of the input data which causes the man and the Room to cease to operate, and both of them separately enter a dynamical ``null'' state. The man is fully operational and cognitively aware, but with no instructions to do anything, he becomes passive and unresponsive. The combined man-book system is fully operational, but the changed input symbols cannot be processed and the entire system freezes. The man is part of the frozen dynamical state and will experience his own qualia associated with being in such a state---likely he will be bored. However, the man's qualia are in no way related to any qualia which are, or are not, experienced by the Chinese Room.  Whether or not the room experiences anything, it will not complain of boredom. Like humans in a disaster situation, the Chinese Room is unable to process or respond meaningfully to the changed environmental symbols. The characteristics of the Room's dynamical null state suggest that the Room is displaying a typical disaster response and, to outside observers, the Room will appear to be cognitively impaired, dysfunctional, lacking direction, unresponsive, shocked and confused.  Anyone making objective observations of the components within the room, i.e. the man and the book, will be able to make inferences to predict and explain the semantic experiences that might be experienced by the Chinese Room should it be so capable.  The same objectively observable dynamical state characteristics can also be used to predict and explain the subjective semantic experiences of humans in similar disaster situations.  

In our approach, we agree with Searle's claim that syntax is not sufficient for semantics \cite{Searle_1984}.  We contend however, that programs are not just syntax as suggested by the systems reply to the Chinese Room.  We justify these claims by focusing on a new class of human semantic symbols invoked during disasters, in which all prelearned or pre-programmed syntactical programs are uncalled.  In these situations, humans and the Chinese Room will exhibit a similar dynamical ``null'' state whose objectively observed characteristics are reminiscent of the subjective semantic experiences of humans in these situations.  Our goal throughout has been to support the systems response to the Chinese Room argument, that there are non-syntactical and non-preprogrammed dynamical states that can arise unexpectedly within the Chinese Room when the environment radically changes, and these states give insight into human experienced qualia.

\subsection{The Subjective-Objective Explanatory Gap and the Properties of Qualia}

In 1974 Nagel held that ``every subjective phenomenon is essentially connected with a single point of view, and it seems inevitable that an objective, physical theory will abandon that point of view.''  The implication was that no presently available conception gives us a clue how a physical theory of mind can account for the subjective character of experience \cite{Nagel_1974_435}.  Later, in 1983, Levine argued ``that psycho-physical identity statements leave a significant {\em explanatory gap}, and, as a corollary, that we don't have any way of determining exactly which psycho-physical identity statements are true.''  Consequently, ``there seems to be nothing about [a physicalist state F] which makes it naturally `fit' the phenomenal properties of pain, any more that it would fit some other set of phenomenal properties''  \cite{Levine_1983_354}. Searle has raised similar concerns about the inability to bridge the subjective-objective divide \cite{Searle_1984}.  

In our approach, we have argued that a radically changed environment can place a learning machine into a dynamical ``null'' state in which fully functional components become quiescent. We have argued that the characteristics of this dynamical null state predict and explain the cognitive response of humans in similar circumstances, who can be passive, unaware, unresponsive, and cognitively impaired.  We have been explicit throughout that the objectively observable characteristics of the null dynamical state of either the machine or the human can be used to predict and explain the subjective experience of humans in similar disaster response circumstances. Thus, we argue that objectively observable facts about a learning machine's dynamical state can be used to predict and explain the subjective content of humans in similar circumstances. We thus take steps to partially close the subjective-objective divide and the explanatory gap. 

Finally, in arguing for eliminativist materialism, Dennett summarized the normally understood properties of qualia as being ineffable, intrinsic, private, and directly apprehensible \cite{Dennett_1988_17}.  We now consider how our model might explain some of these properties. 

First, subjective experiences are considered ineffable, or hard to describe.  A person experiencing a null dynamical state has impaired functionality and will find it difficult to describe their experiences. Any external observers observing a passive and unaware disaster response victim will have non-null states in their own neural networks.  The more observations made by the external observers, the richer their own semantic experiences and the less they replicate victim. Should the observers wish to experience a null dynamical state, they will have to alter their own environment sufficiently to generate their own null states, and this will unfortunately, impair their ability to describe their experience. Essentially, learning a list of syntactical symbols describing a null dynamical state generates a very different, non-null, dynamical state.  

Second, qualia are considered to be intrinsic, or atomic and unanalyzable.  In our example, the null dynamical state is not made up of a concatenation or superposition of several other states, but rather by the absence of all other dynamical states.  Also, as noted above, being in a null state inactivates analytical capabilities. 

Third, qualia are supposed to be private.  As discussed above, making observations of the characteristics of a null state does not duplicate the experience of running a null state.  Learning about what it is like to run a null state requires running the null state in your own neural system.  Others, observing your null state, would learn some objective facts about your dynamical state, but would not learn about your subjective experiences. 

Finally, qualia are supposed to be directly and immediately apprehensible.  When a cognitive system is running a null state, then many aspects of its operation are degraded, and this degradation immediately affects any remaining functional units, now! These units do not have to read syntactical symbols from registers, and then compare those symbols with lookup tables, and then call a subroutine that mimics certain behaviours.  Indeed, the very presence of a null state precludes these normal computational steps from occurring.  A cognitive system running a null state immediately displays degraded performance, and the associated subjective experience would also be immediate.

\section{Dynamical ``Null'' States in Multifunctional Sequentially Programmed Learning Machines}
\label{sect_learning_machine_null_states}

When faced with radical and unexpected environmental change, humans can become cognitively quiescent---apathetic, passive, non-responsive, forgetful, and unable to process information.  We model this response using learning machines which are unable to process data from a changed environment and likely generate null dynamical states.  We will now examine different ways to generate null dynamical states in neural network learning machines in the first part of this section. In the second part of this section, we enrich our model by examining how neural networks can learn to implement multiple functions in simple programmed sequences as specified by environmental data (just as humans do). This will effectively require the neural network to develop its own internal computer architecture from scratch, and it will naturally incorporate environmental data into its naming and addressing conventions, and into its control-flow and sequencing commands. This approach is useful as humans likewise incorporate environmental information into their internal computer architecture and their simple programmed and sequenced actions.  By modeling how a network's programming depends on environmental factors, we are better able to model a more sophisticated range of human responses to environmental change.

\subsection{Neural Network Null States}

We are interested in the dynamical state response of neural networks to environmental change.  Many textbooks on artificial neural networks \cite{Rojas_1996,Haykin_1999,Hagan_2017} and on deep learning networks \cite{Nielsen_2015,Goodfellow_2016} will implicitly include a dynamical equation governing a semi-recurrent feed forward perceptron network as something like
\begin{eqnarray}
   a^L_i &=& \phi (W^L_{ij} a^{L-1}_j)   \nonumber \\
         &=& \phi [W^L_{ij} \phi (W^{L-1}_{jk} a^{L-2}_k)]   \nonumber \\
         &=& \phi \left[W^L_{ij} \phi \left(W^{L-1}_{jk} \dots 
				           \phi [W^3_{rs} \phi (W^2_{st} a^1_t)] \right) \right].
\label{eq:network_dynamical_equation}
\end{eqnarray}  
Here, we show a network of $L$ layers with layer $1$ taking input from the environment, and also possibly recurrently from the $l^{\rm th}$ layer with $2\leq l\leq L$.  We assume the $l^{\rm th}$ layer contains $n_l$ neurons.  The $i^{\rm th}$ neuron of layer $l$, with output $a^l_i$, will take information from the $a^{l-1}_j$ neuron of the previous layer weighted by values $W^l_{ij}$. The input to neuron $a^l_i$ is then $W^l_{ij} a^{l-1}_j$ where we assume an implied summation over repeated indices.  As usual, if the neuron's total input exceeds its bias value, $W^l_{ij} a^{l-1}_j>b^l_i$, the neuron will fire. The bias values can be folded into the weight matrices by the usual process of adding an extra weight into each neuron equal to the negative bias $W^l_{i(n_l+1)}=-b^l_i$ and assuming an additional neuron in every layer that is always active, $a^l_{n_l +1} = 1$. In general, we desire our neurons to be quiescent in the absence of input (though alternative choices could be made, as when a self-initiating oscillatory circuit is required).  This can be achieved by setting all bias values to be positive, so that in the absence of input each neuron remains off and quiescent.  The total input to each neuron, $W^l_{ij} a^{l-1}_j=x$, is then fed into a nonlinear sigmoid function 
\begin{equation}
   \phi(x) = \frac{1}{1+e^{-x}},
\label{eq:sigmoid}
\end{equation}
though other functions provide better performance for deep networks. As usual, the neuron fires iff the argument $x>0$.  

It would generally be considered impossible to analytically ``solve'' the above dynamical equation as interesting networks can have hundreds of millions of parameters.  We have argued however that even the largest and most sophisticated of networks can enter a known state, a null dynamical state, when faced with a radically changed environment well outside its previous training set. This novel data will cause the network to act just like any other untrained network faced with unlearned data, and it will produce random noise.  It will be unable to extract information from that novel data to pass to subsequent layers, and in turn, these layers will be unable to extract or process any useful information. Consequently, the network enters a null dynamical state where each subsequent layer of neurons switches off, one after the other.  Irrespective of how complex the network is, and how well it is trained, we can ``solve'' the network dynamics.  We then have a mathematically specified and objectively known dynamical state which predicts and explains the subjective cognitive sensations experienced by people in similar circumstances.  

There are many different ways to generate a null dynamical state in a neural network governed by Eq. \ref{eq:network_dynamical_equation}.  

{\bf Zero Propagation: $a^l_i \approx 0$:} Because the bias values $b^l_i$ are positive, neurons will be quiescent in the absence of input, and zero inputs to any layer will propagate to generate a general dynamical null state in all downstream regions.  In brief, when $a^{l-1}_k$ is a zero vector in some region, then $a^l_j=\phi(W^{l}_{jk} a^{l-1}_k)=\phi(-b^{l}_j)=0$ as $b^l_j>0$.  Humans can experience this dynamical state by closing their eyes and listening to silence. It may seem overly simple, but the first analytic models of human experience should be simple.   

{\bf Untrained Weights: $W^l_{ij} \approx 0$:} When a region of our network is untrained and has weights close to zero, then all layer outputs in that region will be zero, and these zeros will propagate. That is, when $W^l_{ij} \approx 0$ we have $a^l_i=\phi (W^{l}_{ij} a^{l-1}_j) \approx \phi(-b^{l}_i)=0$ as $b^l_j>0$. Humans can generate and experience this state simply by listening to an unknown language. Humans will experience qualia associated with the absence of information processing, and this semantic experience is predicted by the null dynamical state. 

{\bf Null States by Design:} It also seems possible that the human cognitive system will deliberately not process information in a number of situations.  For example, the vision system doesn't process data during eye saccades (rapid eye movements across the field of view), dream experiences fade from memory and are generally unavailable for later processing, and having an attentional spotlight focused on one stream of data means that many other streams are unattended and are zeroed. Further, the cognitive system suffers limitations which impact cognitive processing. For example, humans can fail to process environmental information in the two cases where information either arrives so fast it averages to zero, or so slowly that the many neural layers trained to identify novelty and change cease to generate output.  In either case we have $a^l_i \approx 0$. Humans can experience the first case by playing a fast-paced video game at too high a level, and can experience the second case by becoming bored perhaps.  

{\bf Novel Inputs Outside the Training Set: $a \notin T$:}  Our main focus of interest is how humans and learning machines both fail to process novel inputs from well outside their previous training sets.  Because they cannot extract information from their inputs they generate zero output, and these zeros propagate throughout the network.  The generated null dynamical state predicts the cognitive experiences of humans in these situations.  An example of this process will be given after we examine how our networks can implement multiple functions in simple programmed sequences as specified by environmental information.

\subsection{Multifunctional Sequentially Programmed Networks}

We suppose that two researchers, denoted $A$ for Alice and $B$ for Bob, are each training high capacity, recurrent, deep neural networks with many layers ($>20$) and with each layer consisting of many ($>10^4$) neurons. Each network will eventually be required to learn to implement multiple functions in programmed sequences, and this normally requires a relatively sophisticated computer architecture with control-flow capabilities. Human children take years to learn how to enact programmed sequences of actions.  Like a human child, each network is a sealed black box meaning that Alice and Bob are unable to specify any aspect of the design of their neural network.  For instance, they cannot shrink the size of a particular hidden layer to force a data compression at the cost of slower learning, or conversely, expand hidden layer size to improve learning at the cost of a greater chance of overfitting. The researchers cannot specify the modular structure and connectivity of the network, data encoding formats, or how the network is to sequence and control data flow between modules and functional units.  As a black box, Alice and Bob must face the fact that their network will have many learning difficulties, just as humans do.  These learning difficulties can themselves be used to model human learning disabilities.  Because the required computer architecture is not designed into the hardware of the network, it has to be learned by the network as a software emulation embedded within the functions learned by the network. To add an emulated computer architecture alongside the data being processed requires the addition of identification, sequencing, addressing, looping, and other control tags into each processed data vector.  We illustrate this now.

First, each network is tasked to learn many functions simultaneously.  We suppose the network is embodied with many input channels (left and right and front and rear cameras and microphones) and that the machine has the ability to speak or write its outputs.  Within the machine, all the different input data streams are addressed under software control to many different functional modules for processing and combination.  Each layer of processing interprets the data at a higher and higher level and adds information to the data stream accompanied by control tags to ensure appropriate downstream processing. The embedded control tags within Alice's machine might indicate that a data stream is sourced from reality (denoted by bit pattern $R$), that it was received at the current time (indicated by a bit pattern $T_1$), that it shows Alice (denoted by bits $A$) in her laboratory (bits $L$) holding a card containing symbols (with input bits ``say $+,4,2$'').  In this case, the network will learn to process the input $(R,T_1,A,L,{\rm say},+,4,2)$, and say the answer $6$.  Other example input-output pairs involving four functions might include 
\begin{eqnarray}
{\rm Input}                   &\rightarrow& {\rm Output}   \nonumber \\
R,T_1,A,L,{\rm say},+,4,2     &\rightarrow& {\rm say }   \; 6 \nonumber \\
R,T_1,A,L,{\rm say},-,4,2     &\rightarrow& {\rm say }   \; 2 \nonumber \\
R,T_1,A,L,{\rm write},*,4,2   &\rightarrow& {\rm write } \; 8 \nonumber \\
R,T_1,A,L,{\rm say},/,4,2     &\rightarrow& {\rm say }   \; 2.
\end{eqnarray}
These examples illustrate a straightforward input-output mapping that can readily be implemented in any suitable network.  

Second, the machine will need to learn to implement simple sequential programs. We presume that once the machine has learned the above four functions, Alice will start teaching it other simple mathematical functions, e.g. $n^2$, as well as how to implement simple sequenced operations as specified by input programs, e.g. $1+2*3$ or $1+2*(3-4/2)$. To accomplish such sequential processing, the machine will need to recognize the order of operator precedence and recurrently route the input expression through the machine multiple times in order to complete its evaluation.  To accomplish this, the machine will need to add control instructions to the input data expression to ensure appropriate addressing, looping, and termination procedures. Human children initially find this very difficult, and are initially trained to use paper and pencil to implement these controls. The association between input data and control codes, essentially data packets containing both data and processing and addressing instructions, leads naturally to a packet-switched network architecture.  In hypothesizing a packet switched learning machine as a model for human cognition, we follow Ref. \cite{Graham_2011_267,Graham_2014_44}. 

Third, we presume the network has an associative memory that can efficiently store previous learning sessions and can recall simultaneous multiple sessions on demand to assist with solving current problems. For instance, at a later time ($T_2$) the machine's memory can generate a data stream (tagged by bits $M$) which encodes the earlier real input-output packet via the bit pattern   
\begin{equation}
  [M,T_2,(R,T_1,A,L,{\rm say},+,4,2\rightarrow 6)].
\end{equation}
Further layers of recursive packet embedding might be possible.  At time $T_3$, the machine can access a memory of having a memory at time $T_2$ of an earlier real event at time $T_1$ via
\begin{equation}
  \{M,T_3,[M,T_2,(R,T_1,A,L,{\rm say},+,4,2\rightarrow 6)]\}.
\end{equation}

Fourth, we expect the network to be able to apply a broad range of analytical functions to its input. For instance, the machine might learn to respond to questions like ``Is Alice visible?'' by polling its input vision data streams for packets which are time stamped ``now'' and embed Alice's code $A$. To questions like ``Is Bob present?'' the machine might poll its processed data streams for packets like $(R, T_{\rm now},B)$ and answer ``Yes'' if these packets exist.  To questions like ``Was Alice present as time $T_1$?'' the machine might present a bit string like $(T_1,A)$ to its memory to trigger associations that return packets like $[M,T_2,(R,T_1,A)]$. If such packets exist, the machine might answer ``Yes.''  

\subsection{Specific and General Function Names Learned from the Environment}

As seen above, our machine makes extensive use of data packets containing data, function names, and control tags, though interpreting which bits belong in which category is very fluid.  In the input packets given above, the ``${\rm say}$'' or ``$+$'' bits could be interpreted as either plain input data, or as function names, or as control tags specifying later sequential machine operations.  A number of different viewpoints are possible, and examining these in turn will suggest how we can embed a sequential program within the data packets processed by a neural network, and some of the difficulties which can arise.  

For a first viewpoint, the bit pattern ``${\rm say},+,4,2$'' has a natural interpretation as a multi-component function name ``${\rm say},+$'' being applied to the data ``$4,2$''.  As the function name comes directly from the environment, any environmental change can corrupt learned function names and hence machine operation.  

A second viewpoint takes the single four-input function learned by the network $f({\rm say},+,4,2)$ with four inputs, and applies a Currying operation to reduce dimensionality to two via $f_{{\rm say},+}(4,2)$.  Alternative inputs then generate different reduced functions, with $f({\rm say},-,4,2)$ becoming $f_{{\rm say},-}(4,2)$. As noted in Ref.~\cite{Wiles_1992_325}, the weights of a neural network are normally considered to encode a single function, but when using Currying operators the weights instead encode an ``interpreter'' able to reconfigure the network to implement different operations under the control of the different control codes in the input data, just like any other stored program computer. This viewpoint again makes it evident that environmental change altering the input data packets can disrupt machine processing. 

A third viewpoint sees these control codes as implementing a combinatorial logic combining the multiple functions learned by the network and selecting for example, alternatives like (${\rm say},+$) or (${\rm write},*$). Incorporating a combinatorial logic within function names allows a leap in the computational complexity of a network---a similar idea was used to explain increased eukaryotic multicellular complexity over that of single-celled prokaryotes \cite{Mattick_2005_856,Mattick_2001_1611}.  Again, when a machine's controlling combinatorial logic is sourced from the environment, this makes it natural to expect that environmental change can disrupt machine operations. 

The final fourth viewpoint recognizes that embedding control codes and data within a single data packet introduces a data flow architecture \cite{Sharp_1992,Hwang_2010}, currently posited as one pathway to exascale computing \cite{Silc_2012,Giorgi_2014_976,Trifunovic_15_2}.  The data packets we have seen above $({\rm say},+,4,2)$ combine operands and data, and these packets are actioned or fire when all operands are present and correct. Self-evidently, unexpected environmental change can mean that these operands are absent or incorrect so again the machine will become dysfunctional.  Other approaches to multitasked neural networks include Refs. \cite{Caruana_1997_41,Kaiser_2017,Ruder_2017}.  

The above viewpoints make it clear that data packets combining data and instructions can allow complex sequential programming.  In parsing its input data packets, the machine must learn by itself which input data is relevant or irrelevant to the task at hand.  In making these decisions, there is a spectrum of possible choices that the machine might make, with no one choice being definitively better than others in every circumstance.  In particular, the machine might assume that the learned function should have high generality, wide applicability and high transferability, and can achieve this by having the shortest name possible. Given input $R,T_1,A,L,{\rm say},+,4,2$, the machine learns to ignore inputs $R,T_1,A,L$ to give the shortest possible name to its learned function ``${\rm say},+$'' to achieve greatest generality.  Conversely, the machine could assume that a learned function is highly specific to the current locality and to the current teacher, and achieves this by having the longest name possible. That is, given input $R,T_1,A,L,{\rm say},+,4,2$, the machine might name its learned function $A,L,{\rm say},+$ so this function is activated only when Alice is present in her laboratory. Humans routinely face the same choice.  When learning a new function, human students must likewise choose between generality and specificity, as some learned functions will be highly general while others will be specific to one location and one time and one person.  As there is no general rule, both the learning machine and the human can often choose incorrectly.  Again, this choice can lead to dysfunctionality, and this is the topic of interest in this paper.

\subsection{A Learned Square Function Faces Environmental Change}

Our goal is to examine how our neural network can become dysfunctional when facing environmental change.  We can illustrate this process by considering how Alice might teach the network to implement an $n^2$ function applied to digits 0-9 via inputs like 
\begin{eqnarray}
{\rm Input}                   &\rightarrow& {\rm Output}   \nonumber \\
R,T_1,A,L,{\rm say},n^2,2     &\rightarrow& {\rm say } \;  4 \nonumber \\
R,T_1,A,L,{\rm write},n^2,0   &\rightarrow& {\rm write } \;  0.
\end{eqnarray}
For illustrative purposes, suppose that the bit pattern for $R,T_i,A,L$ is decimal 16 = binary $10000$, the pattern for ${\rm say}$ in both the input and the output is decimal 4 = binary $100$ while ${\rm write}$ is decimal 6 = binary $110$, and the pattern for $n^2$ is decimal 7 = $111$.  Assume the input and output numbers have standard binary encodings.  Altogether, the bit patterns of the mappings to be learned might be something like
\begin{eqnarray}
{\rm Input}                   &\rightarrow& {\rm Output}   \nonumber \\
R,T_i,A,L,{\rm say},n^2,2     &\rightarrow& {\rm say } \;  4 \nonumber \\
10000,100,111,0010            &\rightarrow& 100,00100,
\end{eqnarray}
or
\begin{eqnarray}
{\rm Input}                   &\rightarrow& {\rm Output}   \nonumber \\
R,T_i,A,L,{\rm write},n^2,2     &\rightarrow& {\rm write } \;  4 \nonumber \\
10000,110,111,0010            &\rightarrow& 110,00100.
\end{eqnarray}
Obviously, the most efficient mapping would simply use an identity to map the input verb ``say'' or ``write'' directly to the output, and another function to map the input number $i$ to the output number $i^2$.  However, the network is large and powerful and performing a random walk around weight space so it might not find these efficient maps, and might instead incorporate unnecessary information into its learned mappings.  For instance, the verb mapping might incorporate the $R,T_i,A,L\equiv 10000$ value into its input via 
\begin{eqnarray}
{\rm Input}                     &\rightarrow& {\rm Output}   \nonumber \\
\underbrace{10000,1x0},111,0010 &\rightarrow& 1x0            \nonumber \\
f_{\rm verb}(10000,1x0)         &\rightarrow& 1x0            \nonumber \\
f^{\rm dec}_{\rm verb}(132+2x)  &\rightarrow& 4+2x,
\end{eqnarray}
where $x=0$ for ``say'' and $x=1$ for ``write''. That is, the decimal function $f^{\rm dec}_{\rm verb}(132+2x)$ maps an input of decimal $132+2x\equiv (10000,1x0)$ to the desired output (4+2x).  Similarly, the ``square'' mapping might incorporate the $R,T_i,A,L\equiv 10000$, the $n^2\equiv 111$, and the input digits $0010$ into its input via 
\begin{eqnarray}
{\rm Input}                     &\rightarrow& {\rm Output}   \nonumber \\
\underbrace{10000,111,0010}     &\rightarrow& 00100          \nonumber \\
f_{n^2}(10000,111,0010)         &\rightarrow& 00100          \nonumber \\
f^{\rm dec}_{n^2}(2160+n)       &\rightarrow& n^2.
\end{eqnarray}
Here, the decimal $f^{\rm dec}_{n^2}(k)$ function maps an input of decimal $k=2160+n\equiv (10000,111,0000)+n$ to the desired output $n^2$. Altogether, the machine has effectively learned two mappings equivalent to 
\begin{eqnarray}
f_{\rm verb}(k=132+2x) & = &  k - 128 \;=\; 4+2x  \nonumber \\
f_{n^2}(k=2160+n) & = &  (k - 2160)^2 \;=\; n^2.
\end{eqnarray}

Suppose now that Alice has trained the network and personally tested it within her laboratory and has shown that it is able to say or write square numbers from 0 to 9 with 100\% accuracy.  Unfortunately, these learned mappings fail as soon as the location or the person input strings are changed from $(R,T_i,A,L,{\rm say},n^2,2)$ to $(R,T_i,A',L',{\rm say},n^2,2)$. To illustrate, suppose the original and altered inputs have encodings
\begin{eqnarray}
(R,T_i,A,L,  {\rm say},n^2,2) & \equiv & (10000,100,111,0010)  \nonumber \\
\hspace{-2cm}(R,T_i,A',L',{\rm say},n^2,2) & \equiv & (10101,100,111,0010).
\end{eqnarray}
The altered input $10101,100$ asks the verb function to process $f_{\rm verb}(172)$ while the altered input $10101,111,0010$ asks the square function to process $f_{n^2}(2802)$ in decimal, and these give  
\begin{eqnarray}
f_{\rm verb}(172) & = &  172 - 128 \;=\; 44 \;\equiv\; 101100 \;\neq\; {\rm say} \nonumber \\
f_{n^2}(2802) & = &  (2802 - 2160)^2 \;=\; 642^2 \;\neq\; 4.
\end{eqnarray}
Here, we note that the output verb code is not well formed and will be unable to activate any output mechanism.  When naming and addressing codes are absent or ill formed, then no information can flow to subsequent layers.  When humans and learning machines incorporate environmental information within their learned functions, then environmental change can destroy those learned mappings and drive outputs so far out-of-range that no further functional processes can be called.  The human and the learning machine can then enter a null dynamical state.

\section{Attachment, Grief, and Spoofing the Network}
\label{sect_attachment_grief_spoof}

Our path forward should now be clear---we have suggested that multifunctional learning machines will often learn functions which are specific to particular individuals and places so that unexpected environmental change can destroy previously learned functions and generate dynamical null states.  Consequently, these learning machines will respond by either seeking to repair the environment, by relearning the changed environment, or by spoofing their original network addressing schemes to allow continued operation in the altered environment.  We now seek to model situations where human learning is so dependent on people and places that environmental change causes cognitive dysfunction, leading to attempts to either repair the environment, to retrain to suit the new environment, or to spoof their network.  We turn to model human experiences in cognitive symbols such as transference, attachment, grief, and the human response to grief.

\subsection{Transfer of Location or Context Based Learning}

Students and people often exhibit an inability to transfer learning between different locations or contexts \cite{Mestre_2006,McKeough_2013}.  For example, students might find it difficult to transfer trigonometric techniques learned in a mathematics classroom to a science classroom, and fail to even recognize that trigonometry is the basis of navigation and construction methods used outside the classroom.  

We suppose that Alice has spent some considerable time in her laboratory  training her network to implement many varied functions (denoted $f_j$) using input output mappings like 
\begin{eqnarray}
{\rm Input}                   &\rightarrow& {\rm Output}   \nonumber \\
R,T_i,A,L,{\rm say},f_j,n     &\rightarrow& {\rm say } \;  f_j(n).
\end{eqnarray}
We further suppose that, unbeknownst to Alice, many of these learned functions have incorporated the location code $L$ into their naming and addressing schemes.  However, this never becomes apparent to Alice as all testing occurs within the laboratory where the machine is fully functional.  As soon as Alice takes the machine to a new location though, $L\rightarrow L'$ say, many of the functions $f_j(n)$ could become dysfunctional causing the network to enter something like a dynamical null state.  In trying to understand this, Alice might return the machine to the original laboratory, $L$, where it instantly recovers its full capabilities.  

Our learning machine duplicates the learning transfer difficulties faced by humans when they have improperly and unknowingly incorporated location information into their function names.  Any environmental alteration can then invalidate the calling of functions, and the learning machine can enter a null dynamical state, equivalent to the blank state experienced by humans in these situations. For example, a student can pass a trigonometry test within the classroom, and become completely blank when asked to determine a navigational bearing outside the classroom. The characteristics of the machine's dynamical null state predict and explain the sensations experienced by people in these situations. Further, the model suggests remedies commonly used by educators to enable learning transfer---Alice and other teachers should vary their location (or more generally, their context) during learning to enable students to more broadly apply their learning.  This model duplicates human responses in situations like classrooms, and can be readily generalized to cases like culture shock \cite{Ward_2005} or future shock \cite{Toffler_1971}.

\subsection{Attachment}

Suppose now that Alice realizes she can correct the location dependencies of her network by undertaking a lengthy retraining process involving many new input-output pairs like 
\begin{eqnarray}
{\rm Input}                   &\rightarrow& {\rm Output}   \nonumber \\
R,T_1,A,L,{\rm say},f_j,n     &\rightarrow& {\rm say } \; f_j(n) \nonumber \\
R,T_1,A,L',{\rm say},f_j,n    &\rightarrow& {\rm say } \;  f_j(n).
\end{eqnarray}
By varying locations, $L, L',L''$ \dots, the machine eventually learns that the location code is unrelated to the function name or its domain of applicability.  Unfortunately, this new training set still leaves the machine with function names that remain dependent on the code for Alice $A$.  As a result, when Alice is absent and Bob is conducting tests, it is discovered that the machine's performance depends on Alice being present.  In particular, we suppose that when Alice is present and the machine is fully functional, all of the machine's input packets have leading bits equivalent to $R,T_i,A \equiv 10000$ (after retraining to remove the location dependence) and these bits have been incorporated into many function definitions.  In contrast, when Bob is testing the machine, the input packets have leading bits equivalent to $R,T_i,B \equiv 10110$.  As shown previously, this can lead to inoperative functions.  

Suppose now that 80\% of all the network's learned function names have incorporated the Alice code $A$.  Consequently, when Alice is present all functions are called correctly and produce the correct outputs, while when she is absent, the machine's functionality is reduced to 20\% and the machine mainly generates dynamical null states.  As previously, the machine's learned functions and its hardware are perfectly operational; it is only that the environmental change destroys the learned machine naming and addressing architecture which generates dysfunctional null states.  In some sense, the machine has become attached to Alice.  

To explore this idea further, suppose that Alice has programmed the machine to optimize some score with points being awarded for each correct answer to questions.  Suppose further that the machine has noted that keeping Alice centered in its field of view maximizes its score and it then learns to rotate its cameras to keep Alice centered as she moves around the laboratory.  On those occasions when Alice has asked questions while ducking out of sight behind a partition say, the machine noted its declining performance and learned that activating its wheel motors to keep her in view resulted in an increased score.  The machine might then generalize this and learn to use its wheel motors to follow Alice when she leaves the room in order to increase its score. It might then be reasonable to predict that the machine will seek to follow Alice everywhere she goes to the extent possible in order to maintain its functionality.  The machine has now become overly attached to Alice, and will stalk Alice to the extent possible.  

Our network will experience and feel nothing as it lacks these capacities.  Further, our model leaves out the complex interplay of the different drivers---the sex drive, attraction, and attachment---and the different emotion systems associated with specific hormones and neural structures in humans \cite{Fisher_1998_23}.  However, our machine will act as if it is attached allowing the prediction that, when attached, a machine or human will perform perfectly (easily, readily, fluently, comfortably) while in the presence of their significant other, but will exhibit a null state (blank, dysfunctional, vacant, listless, unmotivated) when their significant other is absent.  It might also be predicted that the machine and the human might only choose to perform certain functions when their ``partner" is present, and will avoid those activities when they are absent.  These predicted behaviours are routinely observed in attached mammals and humans who prefer the company of conspecifics, maintain close body contact, display separation anxiety, attempt to restore close contact after separation, report feelings of comfort and reduced anxiety when in contact with their partner, and have their lives deeply entwined with their partners \cite{Fisher_1998_23}.  

Interestingly, Bowlby's attachment theory \cite{Bowlby_1969_1,Bowlby_1973_2,Bowlby_1980_3} models organisms as being cybernetically controlled with complexity ranging from primitive organisms possessing reflex-like ``fixed action patterns'' to organisms which are behaviourally flexible and able to adapt to changes in their environment.  Bowlby noted however, that this very adaptability exacts a price as more complex organisms can be more easily subverted from optimality \cite{Bretherton_1992_759}. In this paper, it is precisely those networks with sufficient complexity to learn their own naming and addressing architecture that can become dysfunctional because that very same architecture.

\subsection{Loss and Grief}

After Alice and Bob note that the network appears to be attached to Alice, and, knowing the linkages between attachment theory and bereavement theory \cite{Shear_2005_253,Parkes_2013,Hall_2014_7}, they decide to investigate whether the machine might also be used to model grief. To engender this state, Alice hands the machine over to Bob and never again enters the laboratory to teach the machine. 

Suppose again that the learning machine has about 80\% of its functionality dependent on Alice being present and visible, and that the machine becomes largely dysfunctional and non-responsive when Alice is permanently absent.  Bob's anthropomorphized description of the ``grieving'' learning machine might be that it appears vague, vacant, undirected, unmotivated, unable to concentrate, hopeless, and impaired. Based on these observations, Alice and Bob are able to make accurate predictions about the human response to grief and loss.  They might predict that grieving humans could experience numbness, a lack of energy, emptiness, heaviness, disorganization, withdrawal, absentmindedness, forgetfulness, and a lack of concentration; all symptoms noted in Sect. \ref{sect_new_class_cognitive_symbols} in this paper. Our network can exhibit a null dynamical state with characteristics which predict and explain the qualia experienced by grieving humans. 

We finally note that humans were never trained how to grieve---grief arises naturally and spontaneously.  Similarly, our network was never trained or programmed to grieve, and its symptoms likewise appeared naturally and spontaneously.  This is because, we claim, our network accurately models and duplicates human learning mechanisms rather than merely simulating human behaviours.  In turn, our model suggests that as humans learn multiple new functions from a parent or partner, they are learning a naming and addressing architecture which incorporates environmental information about their partners within their internal addressing schemes.

\subsection{Grief Response}

Alice and Bob now decide to see how the machine responds to its altered environment and its ``grieving'' dynamical state, and to see if this response can predict and explain aspects of the human response to grief. Alice and Bob might well reason that a learning machine might seek to restore its functionality in three different ways.  These are to either return the environment to its original state by having Bob replaced by Alice, $R,T_i,B \rightarrow R,T_i,A$, or to retrain the network so all functions operate when Bob is present, or to spoof the altered input packets $R,T_i,B \equiv 10110$ so that they look like the original $R,T_i,A \equiv 10000$ allowing them to be correctly processed by the unchanged machine. This last option merely involves rewriting some bits which should be feasible.  

With regard to the first option, we have noted above that the machine will indeed try to maintain Alice within its environment to the extent possible, perhaps by stalking Alice to maximize its score.  However, if Alice is permanently departed (to simulate death), then the machine cannot avail itself of this option. The researchers might still predict though, based on the machine's ``desire'' for Alice to return that humans suffering grief might also pine for their loved one and desire their return.

With regard to the second retraining option, Alice and Bob are aware that retraining all the inoperative functions of the machine will take just as long as the original training, perhaps many months.  Based on this, they might predict that human recovery from grief might also involve months of slow retraining.  

Finally, the third option does not involve making changes to either the altered environment or the many functions learned by the machine.  Rather, this option seeks to simply spoof the altered input packets from $R,T_i,B \equiv 10110$ to the original $R,T_i,A \equiv 10000$ so they can be processed by the existing machine without further training.  Alice and Bob are very interested in exploring this option as it potentially could save months of work.  To spoof the input packets, Bob needs to provide another source of the input code $A$ to the learning machine. Bob might first try to use photographs of Alice (with bit code $A'$) which perhaps causes the machine to perceive the input bits $R,T_i,A' \equiv 10100$ which is a little closer to the desired $R,T_i,A \equiv 10000$.  We might guess that some more of the machine's functions become operational---rather than being 80\% dysfunctional the machine might be only 60\% dysfunctional.  Based on this, Alice and Bob might predict that grieving humans would find photographs of their departed loved ones comforting because of a partial return of functionality.  

Refusing to give up, the researchers might realize that the network contains one source which can give a high fidelity copy of Alice's bit pattern as perceived by the machine.  The network's memory, asked to recall Alice, provides memory-tagged and time-stamped packets like $[M,T_j,(R,T_i,A)] \equiv [11010(10000)]$ (say). Here we indicate that the memory packet contains an exact copy of the bit string generated by Alice in reality.  It is possible that some of the network's trained functions will be able to access the embedded $A$ bit string and recover their functionality.  This leads to the prediction that grieving humans might also take comfort from remembering their departed loved ones, and that these memories will ease the symptoms of grief and partially restore function.  

Finally, Alice and Bob might consider teaching the network a single new function to spoof its packets to regain functionality.  Most of the functions learned by the network can readily read, write, delete and alter the control tags attached to their input and output data packets.  The new spoofing function simply needs to strip the memory control tags from the memory packets $[M,T_j,(R,T_i,A)] \equiv [11010(10000)]$ to generate new packets $(R,T_i,A) \equiv 10000$. The memory packets encode a memory of Alice in reality while the altered packets indicate that Alice is present in reality.  As soon as these packets are present within the network, functionality will be widely restored and approach 100\% because the machine now perceives Alice to be present in reality. Of course, altering memory tags into reality tags is not something that you want your machine to do often---if this happened routinely then it wouldn't be able to distinguish what is real from what is remembered.  However, when the costs of being dysfunctional are high, and retraining times are long, then it might well be beneficial to occasionally alter a memory tag into a reality tag. 

After the successful spoofing operation and the restoration of functionality, the network might well respond to questions like
\begin{center}
	\begin{tabular}{cc}
	Question	& Response \\
	Is Alice visible in the room? & No \\
	Can you touch Alice?          & No \\
	Who is visible in the room?   & Bob \\
	Is Alice present in the room? & Yes \\
	\end{tabular}
\end{center}
The network now has internal data packets giving evidence Alice is currently present in the room but is invisible and cannot be touched. In effect, the network now believes in the presence of invisible people, i.e. spirits. Having been spoofed, the network's perception of reality now differs from reality. A machine that cannot feel, or think, or experience sensations can develop dynamical states containing information that invisible people exist and can be present in a room, and this dynamical state instantly allows the machine to resume full functionality.  

Alice and Bob are now in the position of being able to predict that grieving humans might also learn new functions to spoof their memory data packets and come to believe in invisible and immaterial people who cannot be seen or heard but who are nonetheless present. When this new function is learned, it might be predicted that a human will achieve an almost instantaneous restoration of functionality, and that this restoration might be taken as ample justification for the validity of the new learned function.

\section{Conclusion}
\label{sect_conclusion}

In this paper, we have modeled a new class of human semantic symbols experienced when humans become dysfunctional on extreme environmental change. Rather than having a fully functional machine merely simulate dysfunctional behaviour, we duplicated the mechanism causing human dysfunctionality. Human children don't spend months learning how to become attached or to grieve, and our machines don't get taught functions called ``attachment'' or ``grief''.  Rather, our neural network uses environmental inputs to learn a computer architecture able to implement multiple functions and simple sequential programs.  Because environmental information is deeply embedded within the learned computer architecture, it is natural that environmental change disrupts machine operation and, we have argued, typically generates a null dynamical state.  The characteristics of this null state can then predict and explain the characteristics of human experiences of environmental change. 

In its simplest representation, our machine learns to apply a function $f$ to environmental inputs $E$ (containing person $A$ and location $L$ information say) to generate some desired output $y$ via
\begin{equation}
 y = f(E).
\label{eq:conclusion_01}
\end{equation}
Unexpected environmental change $E\rightarrow E'$ moves the input well outside the machine's prior training set which disrupts internal operations sufficiently to generate a null dynamical state via
\begin{equation}
 0 = f(E').
\label{eq:conclusion_02}
\end{equation}
We have argued there are three main ways that the machine may recover.  The first is to take physical steps to restore the changed environment back to its original state $E'\rightarrow E$ to regain the machine functionality of Eq. \ref{eq:conclusion_01}.  The second is to spend time teaching the network a new function $f'$ to suit the altered environment via
\begin{equation}
 y = f'(E').
\label{eq:conclusion_03}
\end{equation}
The third approach is to have the machine learn a new internal function $g$ to preprocess and convert the altered environmental input $E'$ back to its original form $g(E')=E$.  This means the internal machine perception of the environment now differs from the actual environment which restores machine operation via something like
\begin{equation}
 y = f(E) = f[g(E')] = f \circ g (E').
\label{eq:conclusion_04}
\end{equation}

Our approach relating a dynamical state rather than syntactical program code to cognitive semantic symbols provides a concrete example of the systems reply to Searle's Chinese Room argument.  In this approach, symbols are not static neuron firing patterns but dynamical states of the entire learning machine.  In this, we hope to have partially captured Hofstadter's idea of the ``active'' dynamical symbols underlying Aunt Hillary's cognition \cite{Hofstadter_79}.  And because the dynamical state is observable and mathematically defined, the gap between objective observations and subjective experience has been partially closed.  Lastly, by recognizing that the learned computer architecture likely exploits packets of information containing both data and control tags, we were able to consider how different changes in the environment might affect the network, and were thus able to develop models of a range of human  cognitive symbols including transference, attachment, grief, and typical responses to grief. We accomplished all this in a small and simple model lacking complex abilities because we mimic humans who are unable to access prior learning when facing environmental change.  

It has long been assumed that because human cognition is complex and difficult to understand, then any successful artificial intelligence approach will likewise be complex and difficult to understand.  This assumption reinforces our prejudices suspiciously well.  However, the success of our extremely simple approach implies that human cognition is simple at its core, and a moments thought suggests this is a more reasonable assumption to make.  Few people could design a set of function mappings to emulate a computer architecture and yet this is what our neural networks must learn in a few short years. Human cognition is likely based on the simplest possible  computer architecture capable of implementing multiple functions in sequential programs.  Needless to say, this architecture will likely be highly error prone, but it is these errors which will model many aspects of human cognition.  

Our modeling approach can retrodict, as a first approximation, the development of human cognitive capacities over evolutionary timescales.  Simple neural networks apply one function to data and have no need for control tags or a combinatorial logic learned from the environment.  Based on this, we would predict that simple animals neither exhibit nor experience attachment or grief.  More complex machines might possess control tags to implement simple sequential programs learned from the environment, and when these control tags incorporate enough environmental information then these machines can display attachment and grief behaviours. This allows the prediction that animals which learn extensively from their families and their environment can exhibit attachment, grief and mourning behaviours. These expectations seem reasonable.  It does indeed appear that simple animals lack complex cognition and show no attachment or grief behaviours, while more complex animals appear to possess a mind \cite{Suarez_1981_175,Reiss_2001_5937,Chang_2015_212,Krupenye_2016_110}.  Chimpanzees appear to exhibit grief and mourning behaviours \cite{vanLeeuwen_2016_914,vanLeeuwen_2017_44091}, and show preliminary indications of the beginning of ritual \cite{Kuhl_2016_22219}.  Further, archeological evidence appears to show that small brained hominins were going to great lengths to bury their dead \cite{Berger_2015_e09560,Dirks_2015_e09561,Berger_2017_e24234,Dirks_2017_e24231}, while showing little indication of complex symbolic cognitive capacities.   

Eventually however, as animals become more complex and offspring learn vast numbers of functions and capacities from their families, then grief might become so debilitating that evolutionary pressures could tip to favour the development of a spoofing ability. We can then predict that highly complex machines and animals might be so affected by grief that it becomes beneficial to learn new spoofing functions to relieve the symptoms of grief. Machines and animals capable of control tag rewriting will come to believe in spirits and will develop a sense of spirituality. This delinking of perceived reality from actual reality might then allow the development of the enriched symbology that is such a feature of human cognition.  It is only when you can rewrite reality that you can be truly creative---an imagined story becomes reality, reality becomes imaginary, actual histories are reimagined, a stone becomes jewelry becomes love, and so on. Many have speculated about how homo sapiens undertook a ``great leap forward'' \cite{Diamond_1997}, a ``cognitive revolution'' \cite{Harari_2014}, and became behaviourally modern \cite{Caspari_2013_355,Sterelny_2014_1} about 100,000-70,000 years ago.  It has been hypothesized that homo sapiens share a ``fictive language'' in which fictional stories---money, religion, political and legal structures---become reality \cite{Harari_2014}.  It is possible that having a machine develop illusions, a mismatch between perception and reality, is a first step in developing the ability to experience qualia \cite{Yampolskiy_2017_1712.04020}.  Here, we model this cognitive leap as stemming from the development of a spoofing ability equivalent to a control tag rewrite architecture in the human cognitive system, as a complement to other approaches \cite{Mithen_1996,Pinker_1997}.


\begin{thebibliography}{135}
\expandafter\ifx\csname natexlab\endcsname\relax\def\natexlab#1{#1}\fi
\expandafter\ifx\csname bibnamefont\endcsname\relax
  \def\bibnamefont#1{#1}\fi
\expandafter\ifx\csname bibfnamefont\endcsname\relax
  \def\bibfnamefont#1{#1}\fi
\expandafter\ifx\csname citenamefont\endcsname\relax
  \def\citenamefont#1{#1}\fi
\expandafter\ifx\csname url\endcsname\relax
  \def\url#1{\texttt{#1}}\fi
\expandafter\ifx\csname urlprefix\endcsname\relax\def\urlprefix{URL }\fi
\providecommand{\bibinfo}[2]{#2}
\providecommand{\eprint}[2][]{\url{#2}}

\bibitem[{\citenamefont{Chalmers}(1995)}]{Chalmers_95_20}
\bibinfo{author}{\bibfnamefont{D.~J.} \bibnamefont{Chalmers}},
  \bibinfo{journal}{Journal of Consciousness Studies}
  \textbf{\bibinfo{volume}{2}}, \bibinfo{pages}{200} (\bibinfo{year}{1995}).

\bibitem[{\citenamefont{Chalmers}(1996)}]{Chalmers_96}
\bibinfo{author}{\bibfnamefont{D.~J.} \bibnamefont{Chalmers}},
  \emph{\bibinfo{title}{The Conscious Mind: In Search of a Fundamental Theory}}
  (\bibinfo{publisher}{Oxford University Press}, \bibinfo{year}{1996}).

\bibitem[{\citenamefont{Turing}(1937)}]{Turing_1937}
\bibinfo{author}{\bibfnamefont{A.~M.} \bibnamefont{Turing}},
  \bibinfo{journal}{Proceedings of the London Mathematical Society}
  \textbf{\bibinfo{volume}{42}}, \bibinfo{pages}{230} (\bibinfo{year}{1937}).

\bibitem[{\citenamefont{Turing}(1950)}]{Turing_1950}
\bibinfo{author}{\bibfnamefont{A.~M.} \bibnamefont{Turing}},
  \bibinfo{journal}{Mind} \textbf{\bibinfo{volume}{49}}, \bibinfo{pages}{433}
  (\bibinfo{year}{1950}).

\bibitem[{\citenamefont{Newell and Simon}(1976)}]{Newell_1976_11}
\bibinfo{author}{\bibfnamefont{A.}~\bibnamefont{Newell}} \bibnamefont{and}
  \bibinfo{author}{\bibfnamefont{H.~A.} \bibnamefont{Simon}},
  \bibinfo{journal}{Communications of the ACM} \textbf{\bibinfo{volume}{19}},
  \bibinfo{pages}{113} (\bibinfo{year}{1976}).

\bibitem[{\citenamefont{Newell}(1988)}]{Newell_88}
\bibinfo{author}{\bibfnamefont{A.}~\bibnamefont{Newell}}, in
  \emph{\bibinfo{booktitle}{The Impact of Herbert A. Simon}}, edited by
  \bibinfo{editor}{\bibfnamefont{D.}~\bibnamefont{Klahr}} \bibnamefont{and}
  \bibinfo{editor}{\bibfnamefont{K.}~\bibnamefont{Kotovsky}}
  (\bibinfo{publisher}{Erlbaum and Associates}, \bibinfo{year}{1988}).

\bibitem[{\citenamefont{Nilsson}(2007)}]{Nilsson_07_9}
\bibinfo{author}{\bibfnamefont{N.~J.} \bibnamefont{Nilsson}}, in
  \emph{\bibinfo{booktitle}{50 Years of Artificial Intelligence: Essays
  Dedicated to the 50th Anniversary of Artificial Intelligence}}, edited by
  \bibinfo{editor}{\bibfnamefont{M.}~\bibnamefont{Lungarella}},
  \bibinfo{editor}{\bibfnamefont{F.}~\bibnamefont{Iida}},
  \bibinfo{editor}{\bibfnamefont{J.}~\bibnamefont{Bongard}}, \bibnamefont{and}
  \bibinfo{editor}{\bibfnamefont{R.}~\bibnamefont{Pfeifer}}
  (\bibinfo{publisher}{Springer}, \bibinfo{year}{2007}), vol.
  \bibinfo{volume}{4850}, pp. \bibinfo{pages}{9--17}.

\bibitem[{\citenamefont{Hofstadter}(1979)}]{Hofstadter_79}
\bibinfo{author}{\bibfnamefont{D.~R.} \bibnamefont{Hofstadter}},
  \emph{\bibinfo{title}{G\"{o}del, Escher, Bach: An Eternal Golden Braid}}
  (\bibinfo{publisher}{Basic Books}, \bibinfo{year}{1979}).

\bibitem[{\citenamefont{Dennett}(1978)}]{Dennett_78_41}
\bibinfo{author}{\bibfnamefont{D.~C.} \bibnamefont{Dennett}},
  \bibinfo{journal}{Synthese} \textbf{\bibinfo{volume}{38}},
  \bibinfo{pages}{415} (\bibinfo{year}{1978}).

\bibitem[{\citenamefont{Block}(1978)}]{Block_78}
\bibinfo{author}{\bibfnamefont{N.}~\bibnamefont{Block}}, in
  \emph{\bibinfo{booktitle}{Perception and Cognition: Issues in the Foundations
  of Psychology}}, edited by \bibinfo{editor}{\bibfnamefont{C.~W.}
  \bibnamefont{Savage}} (\bibinfo{publisher}{University of Minnesota Press},
  \bibinfo{year}{1978}).

\bibitem[{\citenamefont{Searle}(1980)}]{Searle_80_41}
\bibinfo{author}{\bibfnamefont{J.}~\bibnamefont{Searle}},
  \bibinfo{journal}{Behavioral and Brain Sciences}
  \textbf{\bibinfo{volume}{3}}, \bibinfo{pages}{417} (\bibinfo{year}{1980}).

\bibitem[{\citenamefont{Rey}(1997)}]{Rey_97_46}
\bibinfo{author}{\bibfnamefont{G.}~\bibnamefont{Rey}}, in
  \emph{\bibinfo{booktitle}{The Nature of Consciousness}}, edited by
  \bibinfo{editor}{\bibfnamefont{N.}~\bibnamefont{Block}},
  \bibinfo{editor}{\bibfnamefont{O.}~\bibnamefont{Flanagan}}, \bibnamefont{and}
  \bibinfo{editor}{\bibfnamefont{G.}~\bibnamefont{Güzeldere}}
  (\bibinfo{publisher}{MIT Press}, \bibinfo{address}{Cambridge, MA},
  \bibinfo{year}{1997}), pp. \bibinfo{pages}{461--482}.

\bibitem[{\citenamefont{Dennett}(1991)}]{Dennett_91}
\bibinfo{author}{\bibfnamefont{D.~C.} \bibnamefont{Dennett}},
  \emph{\bibinfo{title}{Consciousness Explained}} (\bibinfo{publisher}{Little,
  Brown, and Co}, \bibinfo{address}{Boston}, \bibinfo{year}{1991}).

\bibitem[{\citenamefont{Penrose}(1989)}]{Penrose_89}
\bibinfo{author}{\bibfnamefont{R.}~\bibnamefont{Penrose}},
  \emph{\bibinfo{title}{The Emperor’s New Mind: Computers, Minds and the Laws
  of Physics}} (\bibinfo{publisher}{Oxford University Press},
  \bibinfo{address}{Oxford}, \bibinfo{year}{1989}).

\bibitem[{\citenamefont{Moore}(1990)}]{Moore_90_23}
\bibinfo{author}{\bibfnamefont{C.}~\bibnamefont{Moore}},
  \bibinfo{journal}{Physical Review Letters} \textbf{\bibinfo{volume}{64}},
  \bibinfo{pages}{2354} (\bibinfo{year}{1990}).

\bibitem[{\citenamefont{Chalcraft and Greene}(1994)}]{Chalcraft_94_5}
\bibinfo{author}{\bibfnamefont{A.}~\bibnamefont{Chalcraft}} \bibnamefont{and}
  \bibinfo{author}{\bibfnamefont{M.}~\bibnamefont{Greene}},
  \bibinfo{journal}{Eureka} \textbf{\bibinfo{volume}{53}}, \bibinfo{pages}{5}
  (\bibinfo{year}{1994}).

\bibitem[{\citenamefont{Boden}(1990)}]{Boden_90}
\bibinfo{author}{\bibfnamefont{M.~A.} \bibnamefont{Boden}}, in
  \emph{\bibinfo{booktitle}{The Philosophy of Artificial Intelligence}}, edited
  by \bibinfo{editor}{\bibfnamefont{M.~A.} \bibnamefont{Boden}}
  (\bibinfo{publisher}{Oxford University Press}, \bibinfo{year}{1990}).

\bibitem[{\citenamefont{Harnad}(1990)}]{Harnad_90_33}
\bibinfo{author}{\bibfnamefont{S.}~\bibnamefont{Harnad}},
  \bibinfo{journal}{Physica D} \textbf{\bibinfo{volume}{42}},
  \bibinfo{pages}{335} (\bibinfo{year}{1990}).

\bibitem[{\citenamefont{Taddeo and Floridi}(2005)}]{Taddeo_05_41}
\bibinfo{author}{\bibfnamefont{M.}~\bibnamefont{Taddeo}} \bibnamefont{and}
  \bibinfo{author}{\bibfnamefont{L.}~\bibnamefont{Floridi}},
  \bibinfo{journal}{Journal of Experimental and Theoretical Artificial
  Intelligence} \textbf{\bibinfo{volume}{17}}, \bibinfo{pages}{419}
  (\bibinfo{year}{2005}).

\bibitem[{\citenamefont{Adams et~al.}(2012)\citenamefont{Adams, Arel, Bach,
  Coop, Furlan, Goertzel, Hall, Samsonovich, Scheutz, Schlesinger
  et~al.}}]{Adams_12_25}
\bibinfo{author}{\bibfnamefont{S.~S.} \bibnamefont{Adams}},
  \bibinfo{author}{\bibfnamefont{I.}~\bibnamefont{Arel}},
  \bibinfo{author}{\bibfnamefont{J.}~\bibnamefont{Bach}},
  \bibinfo{author}{\bibfnamefont{R.}~\bibnamefont{Coop}},
  \bibinfo{author}{\bibfnamefont{R.}~\bibnamefont{Furlan}},
  \bibinfo{author}{\bibfnamefont{B.}~\bibnamefont{Goertzel}},
  \bibinfo{author}{\bibfnamefont{J.~S.} \bibnamefont{Hall}},
  \bibinfo{author}{\bibfnamefont{A.}~\bibnamefont{Samsonovich}},
  \bibinfo{author}{\bibfnamefont{M.}~\bibnamefont{Scheutz}},
  \bibinfo{author}{\bibfnamefont{M.}~\bibnamefont{Schlesinger}},
  \bibnamefont{et~al.}, \bibinfo{journal}{AI Magazine}
  \textbf{\bibinfo{volume}{33}}, \bibinfo{pages}{25} (\bibinfo{year}{2012}).

\bibitem[{\citenamefont{Goertzel and Pennachin}(2007)}]{Goertzel_07}
\bibinfo{editor}{\bibfnamefont{B.}~\bibnamefont{Goertzel}} \bibnamefont{and}
  \bibinfo{editor}{\bibfnamefont{C.}~\bibnamefont{Pennachin}}, eds.,
  \emph{\bibinfo{title}{Artificial General Intelligence}}, Cognitive
  Technologies (\bibinfo{publisher}{Springer}, \bibinfo{year}{2007}).

\bibitem[{\citenamefont{Nilsson}(2005)}]{Nilsson_05_68}
\bibinfo{author}{\bibfnamefont{N.~J.} \bibnamefont{Nilsson}},
  \bibinfo{journal}{AI Magazine} \textbf{\bibinfo{volume}{26}},
  \bibinfo{pages}{68} (\bibinfo{year}{2005}).

\bibitem[{\citenamefont{Minsky et~al.}(2004)\citenamefont{Minsky, Singh, and
  Sloman}}]{Minsky_04_11}
\bibinfo{author}{\bibfnamefont{M.}~\bibnamefont{Minsky}},
  \bibinfo{author}{\bibfnamefont{P.}~\bibnamefont{Singh}}, \bibnamefont{and}
  \bibinfo{author}{\bibfnamefont{A.}~\bibnamefont{Sloman}},
  \bibinfo{journal}{AI Magazine} \textbf{\bibinfo{volume}{25}},
  \bibinfo{pages}{113} (\bibinfo{year}{2004}).

\bibitem[{\citenamefont{Campbell et~al.}(2002)\citenamefont{Campbell, {Hoane
  Jr.}, and Hsu}}]{Campbell_02_57}
\bibinfo{author}{\bibfnamefont{M.}~\bibnamefont{Campbell}},
  \bibinfo{author}{\bibfnamefont{A.~J.} \bibnamefont{{Hoane Jr.}}},
  \bibnamefont{and} \bibinfo{author}{\bibfnamefont{F.}~\bibnamefont{Hsu}},
  \bibinfo{journal}{Artificial Intelligence} \textbf{\bibinfo{volume}{134}},
  \bibinfo{pages}{57} (\bibinfo{year}{2002}).

\bibitem[{\citenamefont{Ferrucci et~al.}(2010)\citenamefont{Ferrucci, Brown,
  Chu-Carroll, Fan, Gondek, Kalyanpur, Lally, Murdock, Nyberg, Prager
  et~al.}}]{Ferrucci_10_59}
\bibinfo{author}{\bibfnamefont{D.}~\bibnamefont{Ferrucci}},
  \bibinfo{author}{\bibfnamefont{E.}~\bibnamefont{Brown}},
  \bibinfo{author}{\bibfnamefont{J.}~\bibnamefont{Chu-Carroll}},
  \bibinfo{author}{\bibfnamefont{J.}~\bibnamefont{Fan}},
  \bibinfo{author}{\bibfnamefont{D.}~\bibnamefont{Gondek}},
  \bibinfo{author}{\bibfnamefont{A.~A.} \bibnamefont{Kalyanpur}},
  \bibinfo{author}{\bibfnamefont{A.}~\bibnamefont{Lally}},
  \bibinfo{author}{\bibfnamefont{J.~W.} \bibnamefont{Murdock}},
  \bibinfo{author}{\bibfnamefont{E.}~\bibnamefont{Nyberg}},
  \bibinfo{author}{\bibfnamefont{J.}~\bibnamefont{Prager}},
  \bibnamefont{et~al.}, \bibinfo{journal}{AI Magazine}
  \textbf{\bibinfo{volume}{31}}, \bibinfo{pages}{59} (\bibinfo{year}{2010}).

\bibitem[{\citenamefont{Hinton et~al.}(2006)\citenamefont{Hinton, Osindero, and
  Teh}}]{Hinton_06_15}
\bibinfo{author}{\bibfnamefont{G.~E.} \bibnamefont{Hinton}},
  \bibinfo{author}{\bibfnamefont{S.}~\bibnamefont{Osindero}}, \bibnamefont{and}
  \bibinfo{author}{\bibfnamefont{Y.}~\bibnamefont{Teh}},
  \bibinfo{journal}{Neural Computation} \textbf{\bibinfo{volume}{18}},
  \bibinfo{pages}{1527} (\bibinfo{year}{2006}).

\bibitem[{\citenamefont{Deng and Yu}(2013)}]{Deng_13_197}
\bibinfo{author}{\bibfnamefont{L.}~\bibnamefont{Deng}} \bibnamefont{and}
  \bibinfo{author}{\bibfnamefont{D.}~\bibnamefont{Yu}},
  \bibinfo{journal}{Foundations and Trends in Signal Processing}
  \textbf{\bibinfo{volume}{7}}, \bibinfo{pages}{197} (\bibinfo{year}{2013}).

\bibitem[{\citenamefont{LeCun et~al.}(2015)\citenamefont{LeCun, Bengio, and
  Hinton}}]{LeCun_15_436}
\bibinfo{author}{\bibfnamefont{Y.}~\bibnamefont{LeCun}},
  \bibinfo{author}{\bibfnamefont{Y.}~\bibnamefont{Bengio}}, \bibnamefont{and}
  \bibinfo{author}{\bibfnamefont{G.}~\bibnamefont{Hinton}},
  \bibinfo{journal}{Nature Review} \textbf{\bibinfo{volume}{521}},
  \bibinfo{pages}{436} (\bibinfo{year}{2015}).

\bibitem[{\citenamefont{Mnih et~al.}(2015)\citenamefont{Mnih, Kavukcuoglu,
  Silver, Rusu, Veness, Bellemare, Graves, Riedmiller, Fidjeland, Ostrovski
  et~al.}}]{Mnih_15_529}
\bibinfo{author}{\bibfnamefont{V.}~\bibnamefont{Mnih}},
  \bibinfo{author}{\bibfnamefont{K.}~\bibnamefont{Kavukcuoglu}},
  \bibinfo{author}{\bibfnamefont{D.}~\bibnamefont{Silver}},
  \bibinfo{author}{\bibfnamefont{A.~A.} \bibnamefont{Rusu}},
  \bibinfo{author}{\bibfnamefont{J.}~\bibnamefont{Veness}},
  \bibinfo{author}{\bibfnamefont{M.~G.} \bibnamefont{Bellemare}},
  \bibinfo{author}{\bibfnamefont{A.}~\bibnamefont{Graves}},
  \bibinfo{author}{\bibfnamefont{M.}~\bibnamefont{Riedmiller}},
  \bibinfo{author}{\bibfnamefont{A.~K.} \bibnamefont{Fidjeland}},
  \bibinfo{author}{\bibfnamefont{G.}~\bibnamefont{Ostrovski}},
  \bibnamefont{et~al.}, \bibinfo{journal}{Nature}
  \textbf{\bibinfo{volume}{518}}, \bibinfo{pages}{529} (\bibinfo{year}{2015}).

\bibitem[{\citenamefont{Silver et~al.}(2016)\citenamefont{Silver, Huang,
  Maddison, Guez, Sifre, van~den Driessche, Schrittwieser, Antonoglou,
  Panneershelvam, Marc~Lanctot et~al.}}]{Silver_16_484}
\bibinfo{author}{\bibfnamefont{D.}~\bibnamefont{Silver}},
  \bibinfo{author}{\bibfnamefont{A.}~\bibnamefont{Huang}},
  \bibinfo{author}{\bibfnamefont{C.~J.} \bibnamefont{Maddison}},
  \bibinfo{author}{\bibfnamefont{A.}~\bibnamefont{Guez}},
  \bibinfo{author}{\bibfnamefont{L.}~\bibnamefont{Sifre}},
  \bibinfo{author}{\bibfnamefont{G.}~\bibnamefont{van~den Driessche}},
  \bibinfo{author}{\bibfnamefont{J.}~\bibnamefont{Schrittwieser}},
  \bibinfo{author}{\bibfnamefont{I.}~\bibnamefont{Antonoglou}},
  \bibinfo{author}{\bibfnamefont{V.}~\bibnamefont{Panneershelvam}},
  \bibinfo{author}{\bibfnamefont{S.~D.} \bibnamefont{Marc~Lanctot}},
  \bibnamefont{et~al.}, \bibinfo{journal}{Nature}
  \textbf{\bibinfo{volume}{529}}, \bibinfo{pages}{484} (\bibinfo{year}{2016}).

\bibitem[{\citenamefont{Silver et~al.}(2017{\natexlab{a}})\citenamefont{Silver,
  Schrittwieser, Simonyan, Antonoglou, Huang, Guez, Hubert, Baker, Lai, Bolton
  et~al.}}]{Silver_2017_354}
\bibinfo{author}{\bibfnamefont{D.}~\bibnamefont{Silver}},
  \bibinfo{author}{\bibfnamefont{J.}~\bibnamefont{Schrittwieser}},
  \bibinfo{author}{\bibfnamefont{K.}~\bibnamefont{Simonyan}},
  \bibinfo{author}{\bibfnamefont{I.}~\bibnamefont{Antonoglou}},
  \bibinfo{author}{\bibfnamefont{A.}~\bibnamefont{Huang}},
  \bibinfo{author}{\bibfnamefont{A.}~\bibnamefont{Guez}},
  \bibinfo{author}{\bibfnamefont{T.}~\bibnamefont{Hubert}},
  \bibinfo{author}{\bibfnamefont{L.}~\bibnamefont{Baker}},
  \bibinfo{author}{\bibfnamefont{M.}~\bibnamefont{Lai}},
  \bibinfo{author}{\bibfnamefont{A.}~\bibnamefont{Bolton}},
  \bibnamefont{et~al.}, \bibinfo{journal}{Nature}
  \textbf{\bibinfo{volume}{550}}, \bibinfo{pages}{354}
  (\bibinfo{year}{2017}{\natexlab{a}}).

\bibitem[{\citenamefont{Silver et~al.}(2017{\natexlab{b}})\citenamefont{Silver,
  Hubert, Schrittwieser, Antonoglou, Lai, Guez, Lanctot, Sifre, Kumaran,
  Graepel et~al.}}]{Silver_2017_01815}
\bibinfo{author}{\bibfnamefont{D.}~\bibnamefont{Silver}},
  \bibinfo{author}{\bibfnamefont{T.}~\bibnamefont{Hubert}},
  \bibinfo{author}{\bibfnamefont{J.}~\bibnamefont{Schrittwieser}},
  \bibinfo{author}{\bibfnamefont{I.}~\bibnamefont{Antonoglou}},
  \bibinfo{author}{\bibfnamefont{M.}~\bibnamefont{Lai}},
  \bibinfo{author}{\bibfnamefont{A.}~\bibnamefont{Guez}},
  \bibinfo{author}{\bibfnamefont{M.}~\bibnamefont{Lanctot}},
  \bibinfo{author}{\bibfnamefont{L.}~\bibnamefont{Sifre}},
  \bibinfo{author}{\bibfnamefont{D.}~\bibnamefont{Kumaran}},
  \bibinfo{author}{\bibfnamefont{T.}~\bibnamefont{Graepel}},
  \bibnamefont{et~al.}, \bibinfo{journal}{arXiv preprint} p.
  \bibinfo{pages}{arXiv:1712.01815} (\bibinfo{year}{2017}{\natexlab{b}}),
  \urlprefix\url{https://arxiv.org/pdf/1712.01815.pdf}.

\bibitem[{\citenamefont{Crick}(1994)}]{Crick_1994}
\bibinfo{author}{\bibfnamefont{F.}~\bibnamefont{Crick}},
  \emph{\bibinfo{title}{The Astonishing Hypothesis}}
  (\bibinfo{publisher}{Charles Scribner’s Sons}, \bibinfo{address}{New York},
  \bibinfo{year}{1994}).

\bibitem[{\citenamefont{Koch}(2004)}]{Koch_2004}
\bibinfo{author}{\bibfnamefont{C.}~\bibnamefont{Koch}},
  \emph{\bibinfo{title}{The Quest for Consciousness: A Neurobiological
  Approach}} (\bibinfo{publisher}{Roberts \& Company},
  \bibinfo{address}{Englewood, CO}, \bibinfo{year}{2004}).

\bibitem[{\citenamefont{Hawkins and Blakeslee}(2004)}]{Hawkins_2004}
\bibinfo{author}{\bibfnamefont{J.}~\bibnamefont{Hawkins}} \bibnamefont{and}
  \bibinfo{author}{\bibfnamefont{S.}~\bibnamefont{Blakeslee}},
  \emph{\bibinfo{title}{On Intelligence: How a New Understanding of the Brain
  will Lead to the Creation of Truly Intelligent Machines}}
  (\bibinfo{publisher}{Times Books}, \bibinfo{year}{2004}).

\bibitem[{\citenamefont{Merolla et~al.}(2014)\citenamefont{Merolla, Arthur,
  Alvarez-Icaza, Cassidy, Sawada, Akopyan, Jackson, Imam, Guo, Nakamura
  et~al.}}]{Merolla_14_668}
\bibinfo{author}{\bibfnamefont{P.~A.} \bibnamefont{Merolla}},
  \bibinfo{author}{\bibfnamefont{J.~V.} \bibnamefont{Arthur}},
  \bibinfo{author}{\bibfnamefont{R.}~\bibnamefont{Alvarez-Icaza}},
  \bibinfo{author}{\bibfnamefont{A.~S.} \bibnamefont{Cassidy}},
  \bibinfo{author}{\bibfnamefont{J.}~\bibnamefont{Sawada}},
  \bibinfo{author}{\bibfnamefont{F.}~\bibnamefont{Akopyan}},
  \bibinfo{author}{\bibfnamefont{B.~L.} \bibnamefont{Jackson}},
  \bibinfo{author}{\bibfnamefont{N.}~\bibnamefont{Imam}},
  \bibinfo{author}{\bibfnamefont{C.}~\bibnamefont{Guo}},
  \bibinfo{author}{\bibfnamefont{Y.}~\bibnamefont{Nakamura}},
  \bibnamefont{et~al.}, \bibinfo{journal}{Science}
  \textbf{\bibinfo{volume}{345}}, \bibinfo{pages}{668} (\bibinfo{year}{2014}).

\bibitem[{\citenamefont{Towlson et~al.}(2013)\citenamefont{Towlson, Vertes,
  Ahnert, Schafer, and Bullmore}}]{Towlson_2013_6380}
\bibinfo{author}{\bibfnamefont{E.~K.} \bibnamefont{Towlson}},
  \bibinfo{author}{\bibfnamefont{P.~E.} \bibnamefont{Vertes}},
  \bibinfo{author}{\bibfnamefont{S.~E.} \bibnamefont{Ahnert}},
  \bibinfo{author}{\bibfnamefont{W.~R.} \bibnamefont{Schafer}},
  \bibnamefont{and} \bibinfo{author}{\bibfnamefont{E.~T.}
  \bibnamefont{Bullmore}}, \bibinfo{journal}{The Journal of Neuroscience}
  \textbf{\bibinfo{volume}{33}}, \bibinfo{pages}{6380} (\bibinfo{year}{2013}).

\bibitem[{\citenamefont{Givon and Lazar}(2016)}]{Givon_16_fly}
\bibinfo{author}{\bibfnamefont{L.~E.} \bibnamefont{Givon}} \bibnamefont{and}
  \bibinfo{author}{\bibfnamefont{A.~A.} \bibnamefont{Lazar}},
  \bibinfo{journal}{PLoS One} \textbf{\bibinfo{volume}{11}},
  \bibinfo{pages}{e0146581} (\bibinfo{year}{2016}).

\bibitem[{\citenamefont{Oh et~al.}(2014)\citenamefont{Oh, Harris, Ng, Winslow,
  Cain, Mihalas, Wang, Lau, Kuan, Henry et~al.}}]{Oh_14_207}
\bibinfo{author}{\bibfnamefont{S.~W.} \bibnamefont{Oh}},
  \bibinfo{author}{\bibfnamefont{J.~A.} \bibnamefont{Harris}},
  \bibinfo{author}{\bibfnamefont{L.}~\bibnamefont{Ng}},
  \bibinfo{author}{\bibfnamefont{B.}~\bibnamefont{Winslow}},
  \bibinfo{author}{\bibfnamefont{N.}~\bibnamefont{Cain}},
  \bibinfo{author}{\bibfnamefont{S.}~\bibnamefont{Mihalas}},
  \bibinfo{author}{\bibfnamefont{Q.}~\bibnamefont{Wang}},
  \bibinfo{author}{\bibfnamefont{C.}~\bibnamefont{Lau}},
  \bibinfo{author}{\bibfnamefont{L.}~\bibnamefont{Kuan}},
  \bibinfo{author}{\bibfnamefont{A.~M.} \bibnamefont{Henry}},
  \bibnamefont{et~al.}, \bibinfo{journal}{Nature}
  \textbf{\bibinfo{volume}{508}}, \bibinfo{pages}{207} (\bibinfo{year}{2014}).

\bibitem[{\citenamefont{Nishimoto et~al.}(2011)\citenamefont{Nishimoto, Vu,
  Naselaris, Benjamini, Yu, and Gallant}}]{Nishimoto_11_1641}
\bibinfo{author}{\bibfnamefont{S.}~\bibnamefont{Nishimoto}},
  \bibinfo{author}{\bibfnamefont{A.~T.} \bibnamefont{Vu}},
  \bibinfo{author}{\bibfnamefont{T.}~\bibnamefont{Naselaris}},
  \bibinfo{author}{\bibfnamefont{Y.}~\bibnamefont{Benjamini}},
  \bibinfo{author}{\bibfnamefont{B.}~\bibnamefont{Yu}}, \bibnamefont{and}
  \bibinfo{author}{\bibfnamefont{J.~L.} \bibnamefont{Gallant}},
  \bibinfo{journal}{Current Biology} \textbf{\bibinfo{volume}{21}},
  \bibinfo{pages}{1641} (\bibinfo{year}{2011}).

\bibitem[{\citenamefont{Lindell et~al.}(2006)\citenamefont{Lindell, Perry,
  Prater, and Nicholson}}]{Lindell_2006}
\bibinfo{author}{\bibfnamefont{M.~K.} \bibnamefont{Lindell}},
  \bibinfo{author}{\bibfnamefont{R.~W.} \bibnamefont{Perry}},
  \bibinfo{author}{\bibfnamefont{C.}~\bibnamefont{Prater}}, \bibnamefont{and}
  \bibinfo{author}{\bibfnamefont{W.~C.} \bibnamefont{Nicholson}},
  \emph{\bibinfo{title}{Fundamentals of Emergency Management}}
  (\bibinfo{publisher}{Federal Emergency Management Agency (FEMA)},
  \bibinfo{year}{2006}).

\bibitem[{\citenamefont{Wallace}(1956{\natexlab{a}})}]{Wallace_1956_1}
\bibinfo{author}{\bibfnamefont{A.~F.~C.} \bibnamefont{Wallace}},
  \emph{\bibinfo{title}{Human Behavior in Extreme Situations: A Survey of the
  Literature and Suggestions for further Research. Distaster Study Number 1.}},
  National Research Council (U.S.) (\bibinfo{publisher}{Committee for Disaster
  Studies, National Academy of Sciences, National Research Council},
  \bibinfo{year}{1956}{\natexlab{a}}).

\bibitem[{\citenamefont{Wallace}(1956{\natexlab{b}})}]{Wallace_1956_3}
\bibinfo{author}{\bibfnamefont{A.~F.~C.} \bibnamefont{Wallace}},
  \emph{\bibinfo{title}{Tornado in Worcester: An Exploratory Study in
  Individual and Community Behavior in an Extreme Situation. Distaster Study
  Number 3.}}, National Research Council (U.S.) (\bibinfo{publisher}{Committee
  for Disaster Studies, National Academy of Sciences, National Research
  Council}, \bibinfo{year}{1956}{\natexlab{b}}).

\bibitem[{\citenamefont{Wallace and Grumet}(2003)}]{Wallace_2003}
\bibinfo{author}{\bibfnamefont{A.~F.~C.} \bibnamefont{Wallace}}
  \bibnamefont{and} \bibinfo{author}{\bibfnamefont{R.~S.}
  \bibnamefont{Grumet}}, \emph{\bibinfo{title}{Revitalizations and Mazeways}},
  Essays on culture change / Anthony F. C. Wallace. Ed. by Robert S. Grumet
  (\bibinfo{publisher}{University of Nebraska Press}, \bibinfo{year}{2003}).

\bibitem[{\citenamefont{Valent}(2000)}]{Valent_2000_706}
\bibinfo{author}{\bibfnamefont{P.}~\bibnamefont{Valent}},
  \emph{\bibinfo{title}{Encyclopedia of Stress}} (\bibinfo{publisher}{Elsevier
  Science}, \bibinfo{year}{2000}), vol.~\bibinfo{volume}{1} of
  \emph{\bibinfo{series}{Encyclopedia of Stress: A-D}}, chap.
  \bibinfo{chapter}{Disaster Syndrome}, pp. \bibinfo{pages}{706--708}.

\bibitem[{\citenamefont{Ripley}(2009)}]{Ripley_2009}
\bibinfo{author}{\bibfnamefont{A.}~\bibnamefont{Ripley}},
  \emph{\bibinfo{title}{The Unthinkable: Who Survives When Disaster Strikes -
  and Why}} (\bibinfo{publisher}{Random House}, \bibinfo{year}{2009}).

\bibitem[{\citenamefont{Leach}(2011)}]{Leach_11_26}
\bibinfo{author}{\bibfnamefont{J.}~\bibnamefont{Leach}}, \bibinfo{journal}{The
  Psychologist} \textbf{\bibinfo{volume}{24}}, \bibinfo{pages}{26}
  (\bibinfo{year}{2011}).

\bibitem[{\citenamefont{Cullen}(1990)}]{Cullen_1990}
\bibinfo{author}{\bibfnamefont{W.~D.} \bibnamefont{Cullen}},
  \emph{\bibinfo{title}{The Public Inquiry into the Piper Alpha Disaster}}, CM:
  1310 (\bibinfo{publisher}{H.M.S.O., Department of Energy},
  \bibinfo{year}{1990}).

\bibitem[{\citenamefont{Griffith and Hart}(2002)}]{Griffith_02_10}
\bibinfo{author}{\bibfnamefont{J.~D.} \bibnamefont{Griffith}} \bibnamefont{and}
  \bibinfo{author}{\bibfnamefont{C.~L.} \bibnamefont{Hart}},
  \bibinfo{journal}{Perceptual and Motor Skills} \textbf{\bibinfo{volume}{94}},
  \bibinfo{pages}{1089} (\bibinfo{year}{2002}).

\bibitem[{\citenamefont{Leach}(2004)}]{Leach_04_539}
\bibinfo{author}{\bibfnamefont{J.}~\bibnamefont{Leach}},
  \bibinfo{journal}{Aviation, Space, and Environmental Medicine}
  \textbf{\bibinfo{volume}{75}}, \bibinfo{pages}{539} (\bibinfo{year}{2004}).

\bibitem[{\citenamefont{Leach}(2012)}]{Leach_12_1152}
\bibinfo{author}{\bibfnamefont{J.}~\bibnamefont{Leach}},
  \bibinfo{journal}{Aviation, Space, and Environmental Medicine}
  \textbf{\bibinfo{volume}{83}}, \bibinfo{pages}{1152} (\bibinfo{year}{2012}).

\bibitem[{\citenamefont{Barlow}(2002)}]{Barlow_2002}
\bibinfo{editor}{\bibfnamefont{D.~H.} \bibnamefont{Barlow}}, ed.,
  \emph{\bibinfo{title}{Anxiety and its Disorders: The Nature and Treatment of
  Anxiety and Panic}} (\bibinfo{publisher}{Guilford Press},
  \bibinfo{address}{New York}, \bibinfo{year}{2002}).

\bibitem[{\citenamefont{Baldwin}(2013)}]{Baldwin_2013_1549}
\bibinfo{author}{\bibfnamefont{D.~V.} \bibnamefont{Baldwin}},
  \bibinfo{journal}{Neuroscience and Biobehavioral Reviews}
  \textbf{\bibinfo{volume}{37}}, \bibinfo{pages}{1549} (\bibinfo{year}{2013}).

\bibitem[{\citenamefont{Bracha}(2004)}]{Bracha_2004_679}
\bibinfo{author}{\bibfnamefont{H.~S.} \bibnamefont{Bracha}},
  \bibinfo{journal}{CNS Spectrums} \textbf{\bibinfo{volume}{9}},
  \bibinfo{pages}{679} (\bibinfo{year}{2004}).

\bibitem[{\citenamefont{Schauer and Elbert}(2010)}]{Schauer_2010_109}
\bibinfo{author}{\bibfnamefont{M.}~\bibnamefont{Schauer}} \bibnamefont{and}
  \bibinfo{author}{\bibfnamefont{T.}~\bibnamefont{Elbert}},
  \bibinfo{journal}{Zeitschrift fur Psychologie / Journal of Psychology}
  \textbf{\bibinfo{volume}{218}}, \bibinfo{pages}{109} (\bibinfo{year}{2010}).

\bibitem[{\citenamefont{Heidt et~al.}(2005)\citenamefont{Heidt, Marx, and
  Forsyth}}]{Heidt_2005_1157}
\bibinfo{author}{\bibfnamefont{J.~M.} \bibnamefont{Heidt}},
  \bibinfo{author}{\bibfnamefont{B.~P.} \bibnamefont{Marx}}, \bibnamefont{and}
  \bibinfo{author}{\bibfnamefont{J.~P.} \bibnamefont{Forsyth}},
  \bibinfo{journal}{Behavior Research and Therapy}
  \textbf{\bibinfo{volume}{43}}, \bibinfo{pages}{1157} (\bibinfo{year}{2005}).

\bibitem[{\citenamefont{Schmidt et~al.}(2008)\citenamefont{Schmidt, Richey,
  Zvolensky, and Maner}}]{Schmidt_2008_292}
\bibinfo{author}{\bibfnamefont{N.~B.} \bibnamefont{Schmidt}},
  \bibinfo{author}{\bibfnamefont{J.~A.} \bibnamefont{Richey}},
  \bibinfo{author}{\bibfnamefont{M.~J.} \bibnamefont{Zvolensky}},
  \bibnamefont{and} \bibinfo{author}{\bibfnamefont{J.~K.} \bibnamefont{Maner}},
  \bibinfo{journal}{Journal of Behavior Therapy and Experimental Psychiatry}
  \textbf{\bibinfo{volume}{39}}, \bibinfo{pages}{292} (\bibinfo{year}{2008}).

\bibitem[{\citenamefont{Volchan et~al.}(2011)\citenamefont{Volchan, Souza,
  Franklin, Norte, Rocha-Rego, Oliveira, David, Mendlowicz, Coutinho, Fiszman
  et~al.}}]{Volchan_2011_13}
\bibinfo{author}{\bibfnamefont{E.}~\bibnamefont{Volchan}},
  \bibinfo{author}{\bibfnamefont{G.~G.} \bibnamefont{Souza}},
  \bibinfo{author}{\bibfnamefont{C.~M.} \bibnamefont{Franklin}},
  \bibinfo{author}{\bibfnamefont{C.~E.} \bibnamefont{Norte}},
  \bibinfo{author}{\bibfnamefont{V.}~\bibnamefont{Rocha-Rego}},
  \bibinfo{author}{\bibfnamefont{J.~M.} \bibnamefont{Oliveira}},
  \bibinfo{author}{\bibfnamefont{I.~A.} \bibnamefont{David}},
  \bibinfo{author}{\bibfnamefont{M.~V.} \bibnamefont{Mendlowicz}},
  \bibinfo{author}{\bibfnamefont{E.~S.} \bibnamefont{Coutinho}},
  \bibinfo{author}{\bibfnamefont{A.}~\bibnamefont{Fiszman}},
  \bibnamefont{et~al.}, \bibinfo{journal}{Biological Psychology}
  \textbf{\bibinfo{volume}{88}}, \bibinfo{pages}{13} (\bibinfo{year}{2011}).

\bibitem[{\citenamefont{Clayton}(2000)}]{Clayton_2000_304}
\bibinfo{author}{\bibfnamefont{P.~J.} \bibnamefont{Clayton}},
  \emph{\bibinfo{title}{Encyclopedia of Stress}} (\bibinfo{publisher}{Elsevier
  Science}, \bibinfo{year}{2000}), vol.~\bibinfo{volume}{1}, chap.
  \bibinfo{chapter}{Bereavement}, pp. \bibinfo{pages}{304--311}.

\bibitem[{\citenamefont{Horowitz et~al.}(1997)\citenamefont{Horowitz, Siegel,
  Holen, Bonanno, Milbrath, and Stinson}}]{Horowitz_1997_904}
\bibinfo{author}{\bibfnamefont{M.~J.} \bibnamefont{Horowitz}},
  \bibinfo{author}{\bibfnamefont{B.}~\bibnamefont{Siegel}},
  \bibinfo{author}{\bibfnamefont{A.}~\bibnamefont{Holen}},
  \bibinfo{author}{\bibfnamefont{G.~A.} \bibnamefont{Bonanno}},
  \bibinfo{author}{\bibfnamefont{C.}~\bibnamefont{Milbrath}}, \bibnamefont{and}
  \bibinfo{author}{\bibfnamefont{C.~H.} \bibnamefont{Stinson}},
  \bibinfo{journal}{American Journal of Psychiatry}
  \textbf{\bibinfo{volume}{154}}, \bibinfo{pages}{904} (\bibinfo{year}{1997}).

\bibitem[{\citenamefont{Shear and Shair}(2005)}]{Shear_2005_253}
\bibinfo{author}{\bibfnamefont{K.}~\bibnamefont{Shear}} \bibnamefont{and}
  \bibinfo{author}{\bibfnamefont{H.}~\bibnamefont{Shair}},
  \bibinfo{journal}{American Journal of Psychiatry}
  \textbf{\bibinfo{volume}{154}}, \bibinfo{pages}{253} (\bibinfo{year}{2005}).

\bibitem[{\citenamefont{Klein}(1991)}]{Klein_91_93}
\bibinfo{author}{\bibfnamefont{D.~C.} \bibnamefont{Klein}},
  \bibinfo{journal}{Journal of Primary Prevention}
  \textbf{\bibinfo{volume}{12}}, \bibinfo{pages}{93} (\bibinfo{year}{1991}).

\bibitem[{\citenamefont{Rivera et~al.}(2014)\citenamefont{Rivera, Talone,
  Boesser, Jentsch, and Yeh}}]{Rivera_2014}
\bibinfo{author}{\bibfnamefont{J.}~\bibnamefont{Rivera}},
  \bibinfo{author}{\bibfnamefont{A.~B.} \bibnamefont{Talone}},
  \bibinfo{author}{\bibfnamefont{C.~T.} \bibnamefont{Boesser}},
  \bibinfo{author}{\bibfnamefont{F.}~\bibnamefont{Jentsch}}, \bibnamefont{and}
  \bibinfo{author}{\bibfnamefont{M.}~\bibnamefont{Yeh}}, in
  \emph{\bibinfo{booktitle}{Proceedings of the Human Factors and Ergonomics
  Society 58th Annual Meeting}} (\bibinfo{year}{2014}), pp.
  \bibinfo{pages}{1047--1051}.

\bibitem[{\citenamefont{Reisenzein et~al.}(2006)\citenamefont{Reisenzein,
  Bordgen, Holtbernd, and Matz}}]{Reisenzein_2006_295}
\bibinfo{author}{\bibfnamefont{R.}~\bibnamefont{Reisenzein}},
  \bibinfo{author}{\bibfnamefont{S.}~\bibnamefont{Bordgen}},
  \bibinfo{author}{\bibfnamefont{T.}~\bibnamefont{Holtbernd}},
  \bibnamefont{and} \bibinfo{author}{\bibfnamefont{D.}~\bibnamefont{Matz}},
  \bibinfo{journal}{Journal of Personality and Social Psychology}
  \textbf{\bibinfo{volume}{91}}, \bibinfo{pages}{295} (\bibinfo{year}{2006}).

\bibitem[{\citenamefont{Martin}(2014)}]{Martin_2014}
\bibinfo{author}{\bibfnamefont{W.~L.} \bibnamefont{Martin}}, Ph.D. thesis,
  \bibinfo{school}{Griffith University, Qld., Australia}
  (\bibinfo{year}{2014}).

\bibitem[{\citenamefont{Ritchie}(1999)}]{Ritchie_1999_78}
\bibinfo{author}{\bibfnamefont{G.}~\bibnamefont{Ritchie}}, in
  \emph{\bibinfo{booktitle}{The Proceedings of the AISB Symposium on Creative
  Language}} (\bibinfo{address}{Edinburgh, Scotland}, \bibinfo{year}{1999}),
  pp. \bibinfo{pages}{78--85}.

\bibitem[{\citenamefont{Shultz}(1972)}]{Shultz_1972_456}
\bibinfo{author}{\bibfnamefont{T.~R.} \bibnamefont{Shultz}},
  \bibinfo{journal}{Journal of Experimental Child Psychology}
  \textbf{\bibinfo{volume}{13}}, \bibinfo{pages}{456} (\bibinfo{year}{1972}).

\bibitem[{\citenamefont{Alden et~al.}(2000)\citenamefont{Alden, Mukherjee, and
  Hoyer}}]{Alden_2000_1}
\bibinfo{author}{\bibfnamefont{D.~L.} \bibnamefont{Alden}},
  \bibinfo{author}{\bibfnamefont{A.}~\bibnamefont{Mukherjee}},
  \bibnamefont{and} \bibinfo{author}{\bibfnamefont{W.~D.} \bibnamefont{Hoyer}},
  \bibinfo{journal}{Journal of Advertising} \textbf{\bibinfo{volume}{29}},
  \bibinfo{pages}{1} (\bibinfo{year}{2000}).

\bibitem[{\citenamefont{Katz}(1993)}]{Katz_1993_59}
\bibinfo{author}{\bibfnamefont{B.}~\bibnamefont{Katz}},
  \bibinfo{journal}{Connection Science} \textbf{\bibinfo{volume}{5}},
  \bibinfo{pages}{59} (\bibinfo{year}{1993}).

\bibitem[{\citenamefont{Samson et~al.}(2009)\citenamefont{Samson, Hempelmann,
  Huber, and Zysset}}]{Samson_2009_1023}
\bibinfo{author}{\bibfnamefont{A.~C.} \bibnamefont{Samson}},
  \bibinfo{author}{\bibfnamefont{C.~F.} \bibnamefont{Hempelmann}},
  \bibinfo{author}{\bibfnamefont{O.}~\bibnamefont{Huber}}, \bibnamefont{and}
  \bibinfo{author}{\bibfnamefont{S.}~\bibnamefont{Zysset}},
  \bibinfo{journal}{Neuropsychologia} \textbf{\bibinfo{volume}{47}},
  \bibinfo{pages}{1023} (\bibinfo{year}{2009}).

\bibitem[{\citenamefont{Polimeni and Reiss}(2006)}]{Polimeni_2006_347}
\bibinfo{author}{\bibfnamefont{J.}~\bibnamefont{Polimeni}} \bibnamefont{and}
  \bibinfo{author}{\bibfnamefont{J.~P.} \bibnamefont{Reiss}},
  \bibinfo{journal}{Evolutionary Psychology} \textbf{\bibinfo{volume}{4}},
  \bibinfo{pages}{347} (\bibinfo{year}{2006}).

\bibitem[{\citenamefont{Hurley et~al.}(2011)\citenamefont{Hurley, Dennett, and
  Adams}}]{Hurley_2011}
\bibinfo{author}{\bibfnamefont{M.~M.} \bibnamefont{Hurley}},
  \bibinfo{author}{\bibfnamefont{D.~C.} \bibnamefont{Dennett}},
  \bibnamefont{and} \bibinfo{author}{\bibfnamefont{R.~B.} \bibnamefont{Adams}},
  \emph{\bibinfo{title}{Inside Jokes: Using Humor to Reverse-Engineer the
  Mind}} (\bibinfo{publisher}{MIT Press}, \bibinfo{year}{2011}).

\bibitem[{\citenamefont{Holland}(1992)}]{Holland_1992}
\bibinfo{author}{\bibfnamefont{J.~H.} \bibnamefont{Holland}},
  \emph{\bibinfo{title}{Adaptation in Natural and Artificial Systems}}
  (\bibinfo{publisher}{MIT Press}, \bibinfo{address}{Cambridge, MA.},
  \bibinfo{year}{1992}).

\bibitem[{\citenamefont{Kauffman}(1993)}]{Kauffman_1993}
\bibinfo{author}{\bibfnamefont{S.~A.} \bibnamefont{Kauffman}},
  \emph{\bibinfo{title}{The Origins of Order: Self-Organization and Selection
  in Evolution}} (\bibinfo{publisher}{Oxford University Press},
  \bibinfo{address}{New York}, \bibinfo{year}{1993}).

\bibitem[{\citenamefont{Dennett}(1995)}]{Dennett_1995}
\bibinfo{author}{\bibfnamefont{D.~C.} \bibnamefont{Dennett}},
  \emph{\bibinfo{title}{Darwin's Dangerous Idea: Evolution and the Meanings of
  Life}} (\bibinfo{publisher}{Simon \& Schuster}, \bibinfo{year}{1995}).

\bibitem[{\citenamefont{Nguyen and Widrow}(1990)}]{Nguyen_1990_596}
\bibinfo{author}{\bibfnamefont{D.}~\bibnamefont{Nguyen}} \bibnamefont{and}
  \bibinfo{author}{\bibfnamefont{B.}~\bibnamefont{Widrow}}, in
  \emph{\bibinfo{booktitle}{Proc. SPIE 1293, Applications of Artificial
  Intelligence VIII}} (\bibinfo{year}{1990}), pp. \bibinfo{pages}{596--602}.

\bibitem[{\citenamefont{Jenkins and Yuhas}(1993)}]{Jenkins_1993_718}
\bibinfo{author}{\bibfnamefont{R.~E.} \bibnamefont{Jenkins}} \bibnamefont{and}
  \bibinfo{author}{\bibfnamefont{B.~P.} \bibnamefont{Yuhas}},
  \bibinfo{journal}{IEEE Transactions on Neural Networks}
  \textbf{\bibinfo{volume}{4}}, \bibinfo{pages}{718} (\bibinfo{year}{1993}).

\bibitem[{\citenamefont{Chalmers et~al.}(1992)\citenamefont{Chalmers, French,
  and Hofstadter}}]{Chalmers_1992_185}
\bibinfo{author}{\bibfnamefont{D.~J.} \bibnamefont{Chalmers}},
  \bibinfo{author}{\bibfnamefont{R.~M.} \bibnamefont{French}},
  \bibnamefont{and} \bibinfo{author}{\bibfnamefont{D.~R.}
  \bibnamefont{Hofstadter}}, \bibinfo{journal}{Journal of Experimental and
  Theoretical Artificial Intelligence} \textbf{\bibinfo{volume}{4}},
  \bibinfo{pages}{185} (\bibinfo{year}{1992}).

\bibitem[{\citenamefont{Dordal}(2016)}]{Dordal_2016}
\bibinfo{author}{\bibfnamefont{P.~L.} \bibnamefont{Dordal}},
  \emph{\bibinfo{title}{An Introduction to Computer Networks}}
  (\bibinfo{publisher}{Web Published, Open licence}, \bibinfo{year}{2016}),
  \urlprefix\url{http://intronetworks.cs.luc.edu/}.

\bibitem[{\citenamefont{Watson et~al.}(2014)\citenamefont{Watson, Baker, Bell,
  Gann, Levine, and Losick}}]{Watson_2014}
\bibinfo{author}{\bibfnamefont{J.~D.} \bibnamefont{Watson}},
  \bibinfo{author}{\bibfnamefont{T.~A.} \bibnamefont{Baker}},
  \bibinfo{author}{\bibfnamefont{S.~P.} \bibnamefont{Bell}},
  \bibinfo{author}{\bibfnamefont{A.}~\bibnamefont{Gann}},
  \bibinfo{author}{\bibfnamefont{M.}~\bibnamefont{Levine}}, \bibnamefont{and}
  \bibinfo{author}{\bibfnamefont{R.}~\bibnamefont{Losick}},
  \emph{\bibinfo{title}{Molecular Biology of the Gene}}
  (\bibinfo{publisher}{Pearson}, \bibinfo{year}{2014}).

\bibitem[{\citenamefont{Schofield}(2015)}]{Schofield_2015}
\bibinfo{author}{\bibfnamefont{P.}~\bibnamefont{Schofield}},
  \emph{\bibinfo{title}{Seals and their Context in the Middle Ages}}
  (\bibinfo{publisher}{Oxbow Books}, \bibinfo{year}{2015}).

\bibitem[{\citenamefont{Hector}(1959)}]{Hector_1959}
\bibinfo{author}{\bibfnamefont{L.}~\bibnamefont{Hector}},
  \emph{\bibinfo{title}{Palaeography and Forgery}}, no.~\bibinfo{number}{15} in
  \bibinfo{series}{Borthwick papers} (\bibinfo{publisher}{St. Anthony's Press},
  \bibinfo{year}{1959}).

\bibitem[{\citenamefont{Balloon}(2001)}]{Balloon_2001_905}
\bibinfo{author}{\bibfnamefont{A.~M.} \bibnamefont{Balloon}},
  \bibinfo{journal}{Emory Law Journal} \textbf{\bibinfo{volume}{50}},
  \bibinfo{pages}{905} (\bibinfo{year}{2001}).

\bibitem[{\citenamefont{Chang}(2013)}]{Chang_2013}
\bibinfo{author}{\bibfnamefont{J.}~\bibnamefont{Chang}},
  \emph{\bibinfo{title}{Empress Dowager Cixi: The Concubine Who Launched Modern
  China}} (\bibinfo{publisher}{Knopf}, \bibinfo{year}{2013}).

\bibitem[{\citenamefont{Searle}(1984)}]{Searle_1984}
\bibinfo{author}{\bibfnamefont{J.}~\bibnamefont{Searle}},
  \emph{\bibinfo{title}{Minds, Brains and Science}}
  (\bibinfo{publisher}{Harvard University Press}, \bibinfo{address}{Cambridge,
  Ma.}, \bibinfo{year}{1984}).

\bibitem[{\citenamefont{Nagel}(1974)}]{Nagel_1974_435}
\bibinfo{author}{\bibfnamefont{T.}~\bibnamefont{Nagel}}, \bibinfo{journal}{The
  Philosophical Review} \textbf{\bibinfo{volume}{83}}, \bibinfo{pages}{435}
  (\bibinfo{year}{1974}).

\bibitem[{\citenamefont{Levine}(1983)}]{Levine_1983_354}
\bibinfo{author}{\bibfnamefont{J.}~\bibnamefont{Levine}},
  \bibinfo{journal}{Pacific Philosophical Quarterly}
  \textbf{\bibinfo{volume}{64}}, \bibinfo{pages}{354} (\bibinfo{year}{1983}).

\bibitem[{\citenamefont{Dennett}(1988)}]{Dennett_1988_17}
\bibinfo{author}{\bibfnamefont{D.~C.} \bibnamefont{Dennett}}, in
  \emph{\bibinfo{booktitle}{Consciousness in Modern Science}}, edited by
  \bibinfo{editor}{\bibfnamefont{A.}~\bibnamefont{Marcel}} \bibnamefont{and}
  \bibinfo{editor}{\bibfnamefont{E.}~\bibnamefont{Bisiach}}
  (\bibinfo{publisher}{Oxford University Press}, \bibinfo{year}{1988}).

\bibitem[{\citenamefont{Rojas}(1996)}]{Rojas_1996}
\bibinfo{author}{\bibfnamefont{R.}~\bibnamefont{Rojas}},
  \emph{\bibinfo{title}{Neural Networks: A Systematic Introduction}}
  (\bibinfo{publisher}{Springer}, \bibinfo{address}{Berlin},
  \bibinfo{year}{1996}).

\bibitem[{\citenamefont{Haykin}(1999)}]{Haykin_1999}
\bibinfo{author}{\bibfnamefont{S.}~\bibnamefont{Haykin}},
  \emph{\bibinfo{title}{Neural Networks: A Comprehensive Foundation}}
  (\bibinfo{publisher}{Pearson Prentice Hall}, \bibinfo{address}{Singapore},
  \bibinfo{year}{1999}).

\bibitem[{\citenamefont{Hagan et~al.}(2017)\citenamefont{Hagan, Demuth, Beale,
  and {De Jesus}}}]{Hagan_2017}
\bibinfo{author}{\bibfnamefont{M.~T.} \bibnamefont{Hagan}},
  \bibinfo{author}{\bibfnamefont{H.~B.} \bibnamefont{Demuth}},
  \bibinfo{author}{\bibfnamefont{M.~H.} \bibnamefont{Beale}}, \bibnamefont{and}
  \bibinfo{author}{\bibfnamefont{O.}~\bibnamefont{{De Jesus}}},
  \emph{\bibinfo{title}{Neural Network Design (ebook)}} (\bibinfo{publisher}{M.
  T. Hagan}, \bibinfo{year}{2017}), \urlprefix\url{hagan.okstate.edu/nnd.html}.

\bibitem[{\citenamefont{Nielsen}(2015)}]{Nielsen_2015}
\bibinfo{author}{\bibfnamefont{M.~A.} \bibnamefont{Nielsen}},
  \emph{\bibinfo{title}{Neural Networks and Deep Learning}}
  (\bibinfo{publisher}{Determination Press}, \bibinfo{year}{2015}),
  \urlprefix\url{http://neuralnetworksanddeeplearning.com}.

\bibitem[{\citenamefont{Goodfellow et~al.}(2016)\citenamefont{Goodfellow,
  Bengio, and Courville}}]{Goodfellow_2016}
\bibinfo{author}{\bibfnamefont{I.}~\bibnamefont{Goodfellow}},
  \bibinfo{author}{\bibfnamefont{Y.}~\bibnamefont{Bengio}}, \bibnamefont{and}
  \bibinfo{author}{\bibfnamefont{A.}~\bibnamefont{Courville}},
  \emph{\bibinfo{title}{Deep Learning}} (\bibinfo{publisher}{MIT Press},
  \bibinfo{year}{2016}), \urlprefix\url{http://www.deeplearningbook.org}.

\bibitem[{\citenamefont{Graham and Rockmore}(2011)}]{Graham_2011_267}
\bibinfo{author}{\bibfnamefont{D.~J.} \bibnamefont{Graham}} \bibnamefont{and}
  \bibinfo{author}{\bibfnamefont{D.~N.} \bibnamefont{Rockmore}},
  \bibinfo{journal}{Journal of Cognitive Neuroscience}
  \textbf{\bibinfo{volume}{23}}, \bibinfo{pages}{267} (\bibinfo{year}{2011}).

\bibitem[{\citenamefont{Graham}(2014)}]{Graham_2014_44}
\bibinfo{author}{\bibfnamefont{D.~J.} \bibnamefont{Graham}},
  \bibinfo{journal}{Frontiers in Computational Neuroscience}
  \textbf{\bibinfo{volume}{8}}, \bibinfo{pages}{44:1} (\bibinfo{year}{2014}).

\bibitem[{\citenamefont{Wiles and Bloesch}(1992)}]{Wiles_1992_325}
\bibinfo{author}{\bibfnamefont{J.}~\bibnamefont{Wiles}} \bibnamefont{and}
  \bibinfo{author}{\bibfnamefont{A.}~\bibnamefont{Bloesch}}, in
  \emph{\bibinfo{booktitle}{Advances in Neural Information Processing Systems
  4}}, edited by \bibinfo{editor}{\bibfnamefont{S.~J.} \bibnamefont{Hanson}}
  \bibnamefont{and} \bibinfo{editor}{\bibfnamefont{R.~P.}
  \bibnamefont{Lippman}} (\bibinfo{publisher}{Morgan Kaufmann},
  \bibinfo{address}{San Mateo, CA.}, \bibinfo{year}{1992}), pp.
  \bibinfo{pages}{325--332}.

\bibitem[{\citenamefont{Mattick and Gagen}(2005)}]{Mattick_2005_856}
\bibinfo{author}{\bibfnamefont{J.~S.} \bibnamefont{Mattick}} \bibnamefont{and}
  \bibinfo{author}{\bibfnamefont{M.~J.} \bibnamefont{Gagen}},
  \bibinfo{journal}{Science} \textbf{\bibinfo{volume}{307}},
  \bibinfo{pages}{856} (\bibinfo{year}{2005}).

\bibitem[{\citenamefont{Mattick and Gagen}(2001)}]{Mattick_2001_1611}
\bibinfo{author}{\bibfnamefont{J.~S.} \bibnamefont{Mattick}} \bibnamefont{and}
  \bibinfo{author}{\bibfnamefont{M.~J.} \bibnamefont{Gagen}},
  \bibinfo{journal}{Molecular Biology and Evolution}
  \textbf{\bibinfo{volume}{18}}, \bibinfo{pages}{1611} (\bibinfo{year}{2001}).

\bibitem[{\citenamefont{Sharp}(1992)}]{Sharp_1992}
\bibinfo{author}{\bibfnamefont{J.~A.} \bibnamefont{Sharp}},
  \emph{\bibinfo{title}{Data Flow Computing: Theory and Practice}}
  (\bibinfo{publisher}{Intellect Books}, \bibinfo{year}{1992}).

\bibitem[{\citenamefont{Hwang and Jotwani}(2010)}]{Hwang_2010}
\bibinfo{author}{\bibfnamefont{K.}~\bibnamefont{Hwang}} \bibnamefont{and}
  \bibinfo{author}{\bibfnamefont{N.}~\bibnamefont{Jotwani}},
  \emph{\bibinfo{title}{Advanced Computer Architecture: Parallelism,
  Scalability, Programmability}} (\bibinfo{publisher}{Tata McGraw-Hill
  Education}, \bibinfo{year}{2010}).

\bibitem[{\citenamefont{Silc et~al.}(2012)\citenamefont{Silc, Robic, and
  Ungerer}}]{Silc_2012}
\bibinfo{author}{\bibfnamefont{J.}~\bibnamefont{Silc}},
  \bibinfo{author}{\bibfnamefont{B.}~\bibnamefont{Robic}}, \bibnamefont{and}
  \bibinfo{author}{\bibfnamefont{T.}~\bibnamefont{Ungerer}},
  \emph{\bibinfo{title}{Processor Architecture: From data flow to Superscalar
  and Beyond}} (\bibinfo{publisher}{Springer Science and Business Media},
  \bibinfo{year}{2012}).

\bibitem[{\citenamefont{Giorgi et~al.}(2014)\citenamefont{Giorgi, Badia, Bodin,
  Cohen, Evripidou, Faraboschi, Fechner, Gao, Garbade, Gayatri
  et~al.}}]{Giorgi_2014_976}
\bibinfo{author}{\bibfnamefont{R.}~\bibnamefont{Giorgi}},
  \bibinfo{author}{\bibfnamefont{R.~M.} \bibnamefont{Badia}},
  \bibinfo{author}{\bibfnamefont{F.}~\bibnamefont{Bodin}},
  \bibinfo{author}{\bibfnamefont{A.}~\bibnamefont{Cohen}},
  \bibinfo{author}{\bibfnamefont{P.}~\bibnamefont{Evripidou}},
  \bibinfo{author}{\bibfnamefont{P.}~\bibnamefont{Faraboschi}},
  \bibinfo{author}{\bibfnamefont{B.}~\bibnamefont{Fechner}},
  \bibinfo{author}{\bibfnamefont{G.~R.} \bibnamefont{Gao}},
  \bibinfo{author}{\bibfnamefont{A.}~\bibnamefont{Garbade}},
  \bibinfo{author}{\bibfnamefont{R.}~\bibnamefont{Gayatri}},
  \bibnamefont{et~al.}, \bibinfo{journal}{Microprocessors and Microsystems:
  Embedded Hardware Design} \textbf{\bibinfo{volume}{38}}, \bibinfo{pages}{976}
  (\bibinfo{year}{2014}).

\bibitem[{\citenamefont{Trifunovic et~al.}(2015)\citenamefont{Trifunovic,
  Milutinovic, Salom, and Kos}}]{Trifunovic_15_2}
\bibinfo{author}{\bibfnamefont{N.}~\bibnamefont{Trifunovic}},
  \bibinfo{author}{\bibfnamefont{V.}~\bibnamefont{Milutinovic}},
  \bibinfo{author}{\bibfnamefont{J.}~\bibnamefont{Salom}}, \bibnamefont{and}
  \bibinfo{author}{\bibfnamefont{A.}~\bibnamefont{Kos}},
  \bibinfo{journal}{Journal of Big Data} \textbf{\bibinfo{volume}{2}}
  (\bibinfo{year}{2015}).

\bibitem[{\citenamefont{Caruana}(1997)}]{Caruana_1997_41}
\bibinfo{author}{\bibfnamefont{R.}~\bibnamefont{Caruana}},
  \bibinfo{journal}{Machine Learning} \textbf{\bibinfo{volume}{28}},
  \bibinfo{pages}{41} (\bibinfo{year}{1997}).

\bibitem[{\citenamefont{Kaiser et~al.}(2017)\citenamefont{Kaiser, Gomez,
  Shazeer, Vaswani, Parmar, Jones, and Uszkoreit}}]{Kaiser_2017}
\bibinfo{author}{\bibfnamefont{L.}~\bibnamefont{Kaiser}},
  \bibinfo{author}{\bibfnamefont{A.~N.} \bibnamefont{Gomez}},
  \bibinfo{author}{\bibfnamefont{N.}~\bibnamefont{Shazeer}},
  \bibinfo{author}{\bibfnamefont{A.}~\bibnamefont{Vaswani}},
  \bibinfo{author}{\bibfnamefont{N.}~\bibnamefont{Parmar}},
  \bibinfo{author}{\bibfnamefont{L.}~\bibnamefont{Jones}}, \bibnamefont{and}
  \bibinfo{author}{\bibfnamefont{J.}~\bibnamefont{Uszkoreit}},
  \bibinfo{journal}{arXiv:1706.05137 [cs.LG]}  (\bibinfo{year}{2017}).

\bibitem[{\citenamefont{Ruder}(2017)}]{Ruder_2017}
\bibinfo{author}{\bibfnamefont{S.}~\bibnamefont{Ruder}},
  \bibinfo{journal}{arXiv:1706.05098 [cs.LG]}  (\bibinfo{year}{2017}).

\bibitem[{\citenamefont{Mestre}(2005)}]{Mestre_2006}
\bibinfo{editor}{\bibfnamefont{J.}~\bibnamefont{Mestre}}, ed.,
  \emph{\bibinfo{title}{Transfer of Learning from a Modern Multidisciplinary
  Perspective}}, Current Perspectives on Cognition, Learning and Instruction
  (\bibinfo{publisher}{Information Age Publishing, Incorporated},
  \bibinfo{year}{2005}).

\bibitem[{\citenamefont{McKeough et~al.}(2013)\citenamefont{McKeough, Lupart,
  and Marini}}]{McKeough_2013}
\bibinfo{editor}{\bibfnamefont{A.}~\bibnamefont{McKeough}},
  \bibinfo{editor}{\bibfnamefont{J.}~\bibnamefont{Lupart}}, \bibnamefont{and}
  \bibinfo{editor}{\bibfnamefont{A.}~\bibnamefont{Marini}}, eds.,
  \emph{\bibinfo{title}{Teaching for Transfer: Fostering Generalization in
  Learning}} (\bibinfo{publisher}{Routledge}, \bibinfo{year}{2013}).

\bibitem[{\citenamefont{Ward et~al.}(2005)\citenamefont{Ward, Bochner, and
  Furnham}}]{Ward_2005}
\bibinfo{author}{\bibfnamefont{C.}~\bibnamefont{Ward}},
  \bibinfo{author}{\bibfnamefont{S.}~\bibnamefont{Bochner}}, \bibnamefont{and}
  \bibinfo{author}{\bibfnamefont{A.}~\bibnamefont{Furnham}},
  \emph{\bibinfo{title}{The Psychology of Culture Shock}}
  (\bibinfo{publisher}{Routledge}, \bibinfo{year}{2005}).

\bibitem[{\citenamefont{Toffler}(1971)}]{Toffler_1971}
\bibinfo{author}{\bibfnamefont{A.}~\bibnamefont{Toffler}},
  \emph{\bibinfo{title}{Future Shock}} (\bibinfo{publisher}{Bantam Books},
  \bibinfo{year}{1971}).

\bibitem[{\citenamefont{Fisher}(1998)}]{Fisher_1998_23}
\bibinfo{author}{\bibfnamefont{H.~E.} \bibnamefont{Fisher}},
  \bibinfo{journal}{Human Nature} \textbf{\bibinfo{volume}{9}},
  \bibinfo{pages}{23} (\bibinfo{year}{1998}).

\bibitem[{\citenamefont{Bowlby}(1969)}]{Bowlby_1969_1}
\bibinfo{author}{\bibfnamefont{J.}~\bibnamefont{Bowlby}},
  \emph{\bibinfo{title}{Attachment and Loss, Vol. 1: Attachment}}
  (\bibinfo{publisher}{Basic Books}, \bibinfo{year}{1969}).

\bibitem[{\citenamefont{Bowlby}(1973)}]{Bowlby_1973_2}
\bibinfo{author}{\bibfnamefont{J.}~\bibnamefont{Bowlby}},
  \emph{\bibinfo{title}{Attachment and Loss, Vol. 2: Separation}}
  (\bibinfo{publisher}{Basic Books}, \bibinfo{year}{1973}).

\bibitem[{\citenamefont{Bowlby}(1980)}]{Bowlby_1980_3}
\bibinfo{author}{\bibfnamefont{J.}~\bibnamefont{Bowlby}},
  \emph{\bibinfo{title}{Attachment and Loss, Vol. 3: Loss, Sadness and
  Depression}} (\bibinfo{publisher}{Basic Books}, \bibinfo{year}{1980}).

\bibitem[{\citenamefont{Bretherton}(1992)}]{Bretherton_1992_759}
\bibinfo{author}{\bibfnamefont{I.}~\bibnamefont{Bretherton}},
  \bibinfo{journal}{Developmental Psychology} \textbf{\bibinfo{volume}{28}},
  \bibinfo{pages}{759} (\bibinfo{year}{1992}).

\bibitem[{\citenamefont{Parkes and Prigerson}(2013)}]{Parkes_2013}
\bibinfo{author}{\bibfnamefont{C.~M.} \bibnamefont{Parkes}} \bibnamefont{and}
  \bibinfo{author}{\bibfnamefont{H.~G.} \bibnamefont{Prigerson}},
  \emph{\bibinfo{title}{Bereavement: Studies of Grief in Adult Life}}
  (\bibinfo{publisher}{Routledge}, \bibinfo{year}{2013}).

\bibitem[{\citenamefont{Hall}(2014)}]{Hall_2014_7}
\bibinfo{author}{\bibfnamefont{C.}~\bibnamefont{Hall}},
  \bibinfo{journal}{Bereavement Care} \textbf{\bibinfo{volume}{33}},
  \bibinfo{pages}{7} (\bibinfo{year}{2014}).

\bibitem[{\citenamefont{Suarez and {Gallup Jr.}}(1981)}]{Suarez_1981_175}
\bibinfo{author}{\bibfnamefont{S.~D.} \bibnamefont{Suarez}} \bibnamefont{and}
  \bibinfo{author}{\bibfnamefont{G.~G.} \bibnamefont{{Gallup Jr.}}},
  \bibinfo{journal}{Journal of Human Evolution} \textbf{\bibinfo{volume}{10}},
  \bibinfo{pages}{175} (\bibinfo{year}{1981}).

\bibitem[{\citenamefont{Reiss and Marino}(2001)}]{Reiss_2001_5937}
\bibinfo{author}{\bibfnamefont{D.}~\bibnamefont{Reiss}} \bibnamefont{and}
  \bibinfo{author}{\bibfnamefont{L.}~\bibnamefont{Marino}},
  \bibinfo{journal}{PNAS} \textbf{\bibinfo{volume}{98}}, \bibinfo{pages}{5937}
  (\bibinfo{year}{2001}).

\bibitem[{\citenamefont{Chang et~al.}(2015)\citenamefont{Chang, Fang, Zhang,
  Poo, and Gong}}]{Chang_2015_212}
\bibinfo{author}{\bibfnamefont{L.}~\bibnamefont{Chang}},
  \bibinfo{author}{\bibfnamefont{Q.}~\bibnamefont{Fang}},
  \bibinfo{author}{\bibfnamefont{S.}~\bibnamefont{Zhang}},
  \bibinfo{author}{\bibfnamefont{M.-M.} \bibnamefont{Poo}}, \bibnamefont{and}
  \bibinfo{author}{\bibfnamefont{N.}~\bibnamefont{Gong}},
  \bibinfo{journal}{Current Biology} \textbf{\bibinfo{volume}{25}},
  \bibinfo{pages}{212} (\bibinfo{year}{2015}).

\bibitem[{\citenamefont{Krupenye et~al.}(2016)\citenamefont{Krupenye, Kano,
  Hirata, Call, and Tomasello}}]{Krupenye_2016_110}
\bibinfo{author}{\bibfnamefont{C.}~\bibnamefont{Krupenye}},
  \bibinfo{author}{\bibfnamefont{F.}~\bibnamefont{Kano}},
  \bibinfo{author}{\bibfnamefont{S.}~\bibnamefont{Hirata}},
  \bibinfo{author}{\bibfnamefont{J.}~\bibnamefont{Call}}, \bibnamefont{and}
  \bibinfo{author}{\bibfnamefont{M.}~\bibnamefont{Tomasello}},
  \bibinfo{journal}{Science} \textbf{\bibinfo{volume}{354}},
  \bibinfo{pages}{110} (\bibinfo{year}{2016}).

\bibitem[{\citenamefont{{van Leeuwen} et~al.}(2016)\citenamefont{{van Leeuwen},
  Mulenga, Bodamer, and Cronin}}]{vanLeeuwen_2016_914}
\bibinfo{author}{\bibfnamefont{E.~J.~C.} \bibnamefont{{van Leeuwen}}},
  \bibinfo{author}{\bibfnamefont{I.~C.} \bibnamefont{Mulenga}},
  \bibinfo{author}{\bibfnamefont{M.~D.} \bibnamefont{Bodamer}},
  \bibnamefont{and} \bibinfo{author}{\bibfnamefont{K.~A.}
  \bibnamefont{Cronin}}, \bibinfo{journal}{American Journal of Primatology}
  \textbf{\bibinfo{volume}{78}}, \bibinfo{pages}{914} (\bibinfo{year}{2016}).

\bibitem[{\citenamefont{{van Leeuwen} et~al.}(2017)\citenamefont{{van Leeuwen},
  Cronin, and Haun}}]{vanLeeuwen_2017_44091}
\bibinfo{author}{\bibfnamefont{E.~J.~C.} \bibnamefont{{van Leeuwen}}},
  \bibinfo{author}{\bibfnamefont{K.~A.} \bibnamefont{Cronin}},
  \bibnamefont{and} \bibinfo{author}{\bibfnamefont{D.~B.~M.}
  \bibnamefont{Haun}}, \bibinfo{journal}{Scientific Reports}
  \textbf{\bibinfo{volume}{13}}, \bibinfo{pages}{44091} (\bibinfo{year}{2017}).

\bibitem[{\citenamefont{Kuhl et~al.}(2016)\citenamefont{Kuhl, Kalan,
  Arandjelovic, Aubert, D’Auvergne, Goedmakers, Jones, Kehoe, Regnaut, Tickle
  et~al.}}]{Kuhl_2016_22219}
\bibinfo{author}{\bibfnamefont{H.~S.} \bibnamefont{Kuhl}},
  \bibinfo{author}{\bibfnamefont{A.~K.} \bibnamefont{Kalan}},
  \bibinfo{author}{\bibfnamefont{M.}~\bibnamefont{Arandjelovic}},
  \bibinfo{author}{\bibfnamefont{F.}~\bibnamefont{Aubert}},
  \bibinfo{author}{\bibfnamefont{L.}~\bibnamefont{D’Auvergne}},
  \bibinfo{author}{\bibfnamefont{A.}~\bibnamefont{Goedmakers}},
  \bibinfo{author}{\bibfnamefont{S.}~\bibnamefont{Jones}},
  \bibinfo{author}{\bibfnamefont{L.}~\bibnamefont{Kehoe}},
  \bibinfo{author}{\bibfnamefont{S.}~\bibnamefont{Regnaut}},
  \bibinfo{author}{\bibfnamefont{A.}~\bibnamefont{Tickle}},
  \bibnamefont{et~al.}, \bibinfo{journal}{Scientific Reports}
  \textbf{\bibinfo{volume}{6}}, \bibinfo{pages}{22219} (\bibinfo{year}{2016}).

\bibitem[{\citenamefont{Berger et~al.}(2015)\citenamefont{Berger, Hawks,
  de~Ruiter, Churchill, Schmid, Delezene, Kivell, Garvin, Williams, DeSilva
  et~al.}}]{Berger_2015_e09560}
\bibinfo{author}{\bibfnamefont{L.~R.} \bibnamefont{Berger}},
  \bibinfo{author}{\bibfnamefont{J.}~\bibnamefont{Hawks}},
  \bibinfo{author}{\bibfnamefont{D.~J.} \bibnamefont{de~Ruiter}},
  \bibinfo{author}{\bibfnamefont{S.~E.} \bibnamefont{Churchill}},
  \bibinfo{author}{\bibfnamefont{P.}~\bibnamefont{Schmid}},
  \bibinfo{author}{\bibfnamefont{L.~K.} \bibnamefont{Delezene}},
  \bibinfo{author}{\bibfnamefont{T.~L.} \bibnamefont{Kivell}},
  \bibinfo{author}{\bibfnamefont{H.~M.} \bibnamefont{Garvin}},
  \bibinfo{author}{\bibfnamefont{S.~A.} \bibnamefont{Williams}},
  \bibinfo{author}{\bibfnamefont{J.~M.} \bibnamefont{DeSilva}},
  \bibnamefont{et~al.}, \bibinfo{journal}{eLife} \textbf{\bibinfo{volume}{4}},
  \bibinfo{pages}{e09560} (\bibinfo{year}{2015}).

\bibitem[{\citenamefont{Dirks et~al.}(2015)\citenamefont{Dirks, Berger,
  Roberts, Kramers, Hawks, Randolph-Quinney, Elliott, Musiba, Churchill,
  de~Ruiter et~al.}}]{Dirks_2015_e09561}
\bibinfo{author}{\bibfnamefont{P.~H.} \bibnamefont{Dirks}},
  \bibinfo{author}{\bibfnamefont{L.~R.} \bibnamefont{Berger}},
  \bibinfo{author}{\bibfnamefont{E.~M.} \bibnamefont{Roberts}},
  \bibinfo{author}{\bibfnamefont{J.~D.} \bibnamefont{Kramers}},
  \bibinfo{author}{\bibfnamefont{J.}~\bibnamefont{Hawks}},
  \bibinfo{author}{\bibfnamefont{P.~S.} \bibnamefont{Randolph-Quinney}},
  \bibinfo{author}{\bibfnamefont{M.}~\bibnamefont{Elliott}},
  \bibinfo{author}{\bibfnamefont{C.~M.} \bibnamefont{Musiba}},
  \bibinfo{author}{\bibfnamefont{S.~E.} \bibnamefont{Churchill}},
  \bibinfo{author}{\bibfnamefont{D.~J.} \bibnamefont{de~Ruiter}},
  \bibnamefont{et~al.}, \bibinfo{journal}{eLife} \textbf{\bibinfo{volume}{4}},
  \bibinfo{pages}{e09561} (\bibinfo{year}{2015}).

\bibitem[{\citenamefont{Berger et~al.}(2017)\citenamefont{Berger, Hawks, Dirks,
  Elliott, and Roberts}}]{Berger_2017_e24234}
\bibinfo{author}{\bibfnamefont{L.~R.} \bibnamefont{Berger}},
  \bibinfo{author}{\bibfnamefont{J.}~\bibnamefont{Hawks}},
  \bibinfo{author}{\bibfnamefont{P.~H.} \bibnamefont{Dirks}},
  \bibinfo{author}{\bibfnamefont{M.}~\bibnamefont{Elliott}}, \bibnamefont{and}
  \bibinfo{author}{\bibfnamefont{E.~M.} \bibnamefont{Roberts}},
  \bibinfo{journal}{eLife} \textbf{\bibinfo{volume}{6}},
  \bibinfo{pages}{e24234} (\bibinfo{year}{2017}).

\bibitem[{\citenamefont{Dirks et~al.}(2017)\citenamefont{Dirks, Roberts,
  Hilbert-Wolf, Kramers, Hawks, Dosseto, Duval, Elliott, Evans, Grün
  et~al.}}]{Dirks_2017_e24231}
\bibinfo{author}{\bibfnamefont{P.~H.} \bibnamefont{Dirks}},
  \bibinfo{author}{\bibfnamefont{E.~M.} \bibnamefont{Roberts}},
  \bibinfo{author}{\bibfnamefont{H.}~\bibnamefont{Hilbert-Wolf}},
  \bibinfo{author}{\bibfnamefont{J.~D.} \bibnamefont{Kramers}},
  \bibinfo{author}{\bibfnamefont{J.}~\bibnamefont{Hawks}},
  \bibinfo{author}{\bibfnamefont{A.}~\bibnamefont{Dosseto}},
  \bibinfo{author}{\bibfnamefont{M.}~\bibnamefont{Duval}},
  \bibinfo{author}{\bibfnamefont{M.}~\bibnamefont{Elliott}},
  \bibinfo{author}{\bibfnamefont{M.}~\bibnamefont{Evans}},
  \bibinfo{author}{\bibfnamefont{R.}~\bibnamefont{Grün}}, \bibnamefont{et~al.},
  \bibinfo{journal}{eLife} \textbf{\bibinfo{volume}{6}},
  \bibinfo{pages}{e24231} (\bibinfo{year}{2017}).

\bibitem[{\citenamefont{Diamond}(1997)}]{Diamond_1997}
\bibinfo{author}{\bibfnamefont{J.}~\bibnamefont{Diamond}},
  \emph{\bibinfo{title}{Guns, Germs, and Steel: The Fates of Human Societies}}
  (\bibinfo{publisher}{Norton}, \bibinfo{address}{New York},
  \bibinfo{year}{1997}).

\bibitem[{\citenamefont{Harari}(2014)}]{Harari_2014}
\bibinfo{author}{\bibfnamefont{Y.~N.} \bibnamefont{Harari}},
  \emph{\bibinfo{title}{Sapiens: A Brief History of Humankind}}
  (\bibinfo{publisher}{Vintage}, \bibinfo{year}{2014}).

\bibitem[{\citenamefont{Caspari and Wolpoff}(2013)}]{Caspari_2013_355}
\bibinfo{author}{\bibfnamefont{R.}~\bibnamefont{Caspari}} \bibnamefont{and}
  \bibinfo{author}{\bibfnamefont{M.~H.} \bibnamefont{Wolpoff}}, in
  \emph{\bibinfo{booktitle}{The Origins of Modern Humans: Biology
  Reconsidered}}, edited by \bibinfo{editor}{\bibfnamefont{F.~H.}
  \bibnamefont{Smith}} \bibnamefont{and}
  \bibinfo{editor}{\bibfnamefont{J.~C.~M.} \bibnamefont{Ahern}}
  (\bibinfo{publisher}{John Wiley \& Sons}, \bibinfo{year}{2013}), pp.
  \bibinfo{pages}{355--391}.

\bibitem[{\citenamefont{Sterelny and Hiscock}(2014)}]{Sterelny_2014_1}
\bibinfo{author}{\bibfnamefont{K.}~\bibnamefont{Sterelny}} \bibnamefont{and}
  \bibinfo{author}{\bibfnamefont{P.}~\bibnamefont{Hiscock}},
  \bibinfo{journal}{Biological Theory} \textbf{\bibinfo{volume}{9}},
  \bibinfo{pages}{1} (\bibinfo{year}{2014}).

\bibitem[{\citenamefont{Yampolskiy}(2017)}]{Yampolskiy_2017_1712.04020}
\bibinfo{author}{\bibfnamefont{R.}~\bibnamefont{Yampolskiy}},
  \bibinfo{journal}{arXiv preprint} p. \bibinfo{pages}{arXiv:1712.04020}
  (\bibinfo{year}{2017}), \urlprefix\url{https://arxiv.org/pdf/1712.04020.pdf}.

\bibitem[{\citenamefont{Mithen}(1996)}]{Mithen_1996}
\bibinfo{author}{\bibfnamefont{S.}~\bibnamefont{Mithen}},
  \emph{\bibinfo{title}{The Prehistory of the Mind: A Search for the Origins of
  Art, Religion and Science}} (\bibinfo{publisher}{Thames and Hudson},
  \bibinfo{address}{London}, \bibinfo{year}{1996}).

\bibitem[{\citenamefont{Pinker}(1997)}]{Pinker_1997}
\bibinfo{author}{\bibfnamefont{S.}~\bibnamefont{Pinker}},
  \emph{\bibinfo{title}{How the Mind Works}} (\bibinfo{publisher}{Norton},
  \bibinfo{year}{1997}).

\end{thebibliography}
\end{document}